\documentclass[twoside,11pt]{article}

\usepackage{pdfsync}
\usepackage{jmlr2e}

\usepackage{amsmath,amsfonts,amssymb} 
\usepackage{graphicx}
\usepackage{algorithmic,algorithm}
\newcommand{\s}{\sigma} % posterior variance
\newcommand{\sn}{\s_{\operatorname{noise}}}
\newcommand{\sbet}{\sqrt{\beta}} % exploration/exploitation balance
\renewcommand{\l}{\lambda}
\newcommand{\bl}{\bar{\l}}
\newcommand{\hl}{\hat{\l}}
\newcommand{\hhl}{\hat{\hl}}
\newcommand{\qs}{q_{s}}
\newcommand{\Cs}{C_{s}}
\newcommand{\As}{A_{s}}

\newcommand{\E}{\mathbb{E}} % Expected value
\newcommand{\av}{\mathbf a}

\newcommand{\xv}{\mathbf x}
\newcommand{\yv}{\mathbf y}

\newcommand{\kv}{\mathbf k}
\newcommand{\ov}{\mathbf 0}

\newcommand{\fv}{\mathbf f}
\newcommand{\uv}{\mathbf u}
\newcommand{\vv}{\mathbf v}
\newcommand{\Iv}{\mathbf 1}
\newcommand{\jv}{\mathbf j}
\newcommand{\BI}{\mathbf I}
\newcommand{\BJ}{\mathbf J}
\newcommand{\BU}{\mathbf U}
\newcommand{\BK}{\mathbf K}

\newcommand{\Xs}{\mathcal X} % Set of arms
\newcommand{\Ss}{\mathcal S} % Set?
\renewcommand{\Re}{\mathbb{R}} % Space of real numbers
\newcommand{\reward}{\operatorname{reward}}
\newcommand{\cov}{\operatorname{cov}}

\newcommand{\tO}{\tilde{O}} % O notation, up to a log factor
\newcommand{\abs}[1]{\left|#1\right|} % absolute value: |.|

\ShortHeadings{Gaussian Process Bandits for Tree Search}{Dorard and Shawe-Taylor}
\firstpageno{1}

\begin{document}

%\mainmatter
\title{Gaussian Process Bandits for Tree Search:\\Theory and Application to Planning in Discounted MDPs}
%\titlerunning{Gaussian Process Bandits for Tree Search}
%\author{Louis Dorard
%\and John Shawe-Taylor}
%\institute{UCL}
\author{\name Louis Dorard \email louis@dorard.me
       \AND
       \name John Shawe-Taylor \email jst@cs.ucl.ac.uk \\
       \addr University College London\\ Department of Computer Science\\ London WC1E 6BT, U.K.}

\editor{}

\maketitle

\begin{abstract}
We motivate and analyse a new Tree Search algorithm, GPTS, based on recent theoretical advances in the use of Gaussian Processes for Bandit problems. We consider tree paths as arms and we assume the target/reward function is drawn from a GP distribution. The posterior mean and variance, after observing data, are used to define confidence intervals for the function values, and we sequentially play arms with highest upper confidence bounds.

We give an efficient implementation of GPTS and we adapt previous regret bounds by determining the decay rate of the eigenvalues of the kernel matrix on the whole set of tree paths. We consider two kernels in the feature space of binary vectors indexed by the nodes of the tree: linear and Gaussian. The regret grows in square root of the number of iterations $T$, up to a logarithmic factor, with a constant that improves with bigger Gaussian kernel widths. We focus on practical values of $T$, smaller than the number of arms.

Finally, we apply GPTS to Open Loop Planning in discounted Markov Decision Processes by modelling the reward as a discounted sum of independent Gaussian Processes. We report similar regret bounds to those of the OLOP algorithm.
\end{abstract}


\section{Introduction}
\label{sec:introduction}

In order to motivate the work presented here, we first review the problem of tree search and its bandit-based approaches. We motivate the use of models of arm dependencies in bandit problems, for the purpose of searching trees. We then introduce our approach based on Gaussian Processes, that we analyse in the rest of this paper.

% vocabulary: depth-first, best-first, iterative deepening; see http://en.wikipedia.org/wiki/Alpha-beta_pruning#Other_algorithms
% Why stochastic tree search, compared to branch and bound such as in minimax or alpha beta
% see advantages over alpha beta in http://hal.inria.fr/docs/00/12/15/16/PDF/RR-6062.pdf

\subsection{Context}

Tree search consists in looking for an optimal sequence of nodes to select, starting from the root, in order to maximise a reward given when a leaf is reached. We introduce this problem in more detail, we motivate the use of bandit algorithms for tree search and we review existing techniques.

\subsubsection{Tree search}

\paragraph{Applications}
Tree search is important in Artificial Intelligence for Games, where the machine represents possible sequences of moves as a tree and looks ahead for the first move which is most likely to yield a win. Rewards are given by Monte Carlo simulations where we randomly finish the game from the current position and return 1 for a win, 0 otherwise. Tree search can also be used to search for an optimum in a space of sequences of given length, as in sequence labelling. More generally, it can be used to search any topological space for which a tree of coverings is defined, as shown by \citet{Bubeck2009}, where each node corresponds to a region of the space. For instance, if the space to search is a $d$-dimensional hyper-rectangle, the root node of the tree of coverings is the whole hyper-rectangle, and children nodes are defined recursively by splitting the region of the current node in two: each region is a hyper-rectangle and we split in the middle along the longest side.

\paragraph{Planning in Markov Decision Processes}
In MDPs, an agent takes a sequence of actions that take it into a sequence of states, gets rewards from the environment for each action it takes, and aims at maximising its total reward. Alternatively, a simpler objective is to maximise the discounted sum of rewards the agent gets: a discount factor $0 < \gamma < 1$ is given beforehand and a weight of $\gamma^t$ is applied to the reward obtained at time $t$, for all $t$. If a generative model of the MDP is available (i.e. given a state we can determine the actions available from this state and the rewards obtained for each of these actions, without calling the environment), then we can represent the possible sequences of actions as a tree and determine the reward for each path through this tree (as a discounted sum of intermediate rewards). The idea of using bandit algorithms in the search for an optimal action in large state-space MDPs (i.e. planning) was introduced by \citet{Kocsis2006} and also considered by \citet{Chang2007} \footnote{Bandits are also used by \citet{Ortner2010} for closed-loop planning (where the chosen actions depend on the current states) in MDPs with deterministic transitions.}, as an alternative to costly dynamic programming approaches that aim to approximate the optimal value function.

\paragraph{Challenges}
Searching trees with large branching factors can be computationally challenging, as applications to the game of Go have shown. It requires to efficiently select branches to explore based on their estimated potential (i.e. how good the reward can be at leaves of paths going through this node) and the uncertainty in the estimations. Similarly, high depths can be unattainable due to lack of computational time and bad selection of the branches to explore. A tree search algorithm should not waste too much time in exploring sub-optimal branches, while still exploring enough in order not to miss the optimal. Bandit algorithms can be used to guide the selection on nodes in the exploration of the tree, based on knowledge acquired from previous reward samples. However, one must be cautious that the process of selecting the best nodes to explore first doesn't become itself too computationally expensive. In the work of \citet{Gelly2006} on the search of Go game trees, bandit algorithms allow a more efficient exploration of the tree compared to traditional Branch \& Bound approaches (Alpha-Beta).

\subsubsection{Bandit problems}

% Bandit setting
The bandit problem is a simple model of the trade-off between exploration and exploitation. The multi-armed bandit is an analogy with a traditional slot machine, known as a one-armed bandit, but with multiple arms. In the stochastic bandit scenario, the player, after pulling (or `playing') an arm selected from the finite set of arms, receives a reward. It is assumed that the reward obtained when playing arm $i$ is a sample from a distribution $R_{i}$, unknown to the player, and that samples are iid. A stochastic bandit problem is characterised by a set of probability distributions $R_{i, 1 \leq i \leq N}$ \footnote{Non-stochastic bandit problems are also of interest, as well as problems in which the distributions are allowed to change through time \citep[see][for an overview of the different types of bandit problems]{Bubeck2010a}.}.

% Objective, measure of policy performance through regret
\paragraph{Measure of performance}
The objective of the player is to maximise the collected reward sum (or `cumulative reward') through iterative plays of the bandit. The optimal arm selection policy $S^*$, i.e. the policy that yields maximum expected cumulative reward, consists in selecting arm $i^*=\mbox{argmax}_i \{\E R_i\}$ to play at each iteration. The expected cumulative reward of $S^*$ at time $t$ (after $t$ iterations) is $t \E R_{i^*}$. The performance of a policy $S$ is assessed by the analysis of its cumulative regret at time $T$, defined as the difference between the expected cumulative reward of $S^*$ and $S$ at time $T$. 

% Exploration vs exploitation
\paragraph{Exploration vs. exploitation}
A good policy requires to optimally balance the learning of the distributions $R_i$ and the exploitation of arms which have been learnt as having high expected rewards. When the number of arms is finite and smaller than the number of experiments allowed, it is possible to explore all the possible options (arms) a certain number of times, thus building empirical estimates $\hat{\E} R_i$, and exploit the best performing ones. As the number of times we play the same arm $i$ grows, we expect our reward estimate to improve.

% Optimism in the face of uncertainty
\paragraph{Optimism in the face of uncertainty}
A popular strategy for balancing exploration and exploitation consists in applying the so-called ``optimism in the face of uncertainty'' principle: reward estimates and uncertainty estimates are maintained for each arm, such that the probability that the actual mean-reward values are outside of the confidence intervals drops quickly. The arm to be played at each time step is the one for which the upper bound of the confidence interval is the highest. This strategy, as implemented in the UCB algorithm, has been shown by \citet{Auer2002} to achieve optimal regret growth-rate for problems with independent arms: problem-specific upper bound in $O(\log(T))$, and problem-independent upper bound in $O(\sqrt{T})$ \footnote{A regret bound is said to be problem-specific when it involves quantities that are specific to the current bandit problem, such as the sub-optimalities $\Delta_i = \E R_{i^*} - \E R_{i}$ of arms, based on the means of the distributions for this problem. The second bound, however, does not involve such quantities.}.

\subsubsection{Bandit-based Tree Search algorithms}

Typically, algorithms proceed in iterations. After the $t^{th}$ iteration, a leaf node $n_t$ is selected and a reward $y_t$ is received. It is usually assumed that there exists a mean-reward function $f$ such that $y_t$ is a noisy observation of $f(n_t)$. Other common assumptions are that $f^*$, the highest value of $f$, is known (or an upper bound on $f^*$ is known) and is always bigger than $y_t$. The algorithm stops when a convergence criterion is met, when a computational/time budget is exhausted (in game tree search for instance), or when a maximum number of iterations has been specified (this is referred to as ``fixed horizon'' exploration, by opposition to ``anytime''). In the end, a path through the tree is given. This can simply be the path that leads to the leaf node that received the highest reward.

\paragraph{Path selection as a sequence of bandit problems}
The algorithm developed by \citet{Kocsis2006}, UCT, considers bandit problems at each node in the tree. The children of a given node represent the arms of the associated bandit problem, and the rewards obtained for selecting an arm are the values obtained at a leaf. At each iteration, we start from the root and select children nodes by invoking the bandit algorithms of the parents, until a leaf is reached and a reward is received, which is then back propagated to the ancestors up to the root. The bandit algorithm used in UCT is UCB \citep{Auer2002} which stands for Upper Confidence Bounds and implements the principle of optimism in the face of uncertainty \footnote{In the tree setting however, rewards are not iid and the values used by the UCB algorithms at each node are not true upper confidence bounds.}.

\paragraph{``Smooth'' trees}
Although \citet{Gelly2006} reported that UCT performed very well on Go game trees, it was shown by \citet{Coquelin2007a} that it can behave poorly in certain situations because of ``overly optimistic assumptions in the design of its upper confidence bounds'' \citep{Bubeck2010b}, leading to a high lower bound on its cumulative regret. An other algorithm was proposed, BAST (Bandit Algorithm for Smooth Trees), which can be parameterised to adapt to different levels of smoothness of the reward function on leaves, and to deal with the situations that UCT handles badly. BAST is only different from UCT in the definition of its `upper confidence bounds' (UCT is actually a special case of BAST, corresponding to a particular value of one of the algorithm's parameters). A time-independent regret upper bound was derived, however it was expressed in terms of the sub-optimality values $\Delta_i$ of nodes (dependent on the reward $f$ on nodes, hence unknown to the algorithm) and was thus problem specific. Also, quite paradoxically, the bound could become very high for smooth functions (because of $1/\Delta_i$ terms).

\paragraph{Optimistic planning in discounted MDPs}
The discount factor implies a particular smoothness of the function $f$ on tree paths (the smaller $\gamma$, the smoother the function), which is the starting point of the work of \citet{Bubeck2010b} on the Open Loop Optimistic Planning algorithm, close in spirit to BAST. OLOP has been proved to be minimax optimal, up to a logarithmic factor, which means that the upper bound growth rate of its simple regret \footnote{The simple regret is defined as the difference between $f^*$ and the best value of $f$ for the arms that have been played.} matches the lower bound. However, OLOP requires the knowledge of the time horizon $T$ and the regret bounds do not apply when the algorithm is run in an anytime fashion.

\paragraph{Measure of performance}
A Tree Search algorithm's performance can be measured, as for a bandit algorithm, by its cumulative regret $R_T = T f^* - \sum_{t=1}^T f(n_t)$. However, although this is a good objective to achieve a good exploration/exploitation balance, we might be ultimately interested in a bound on how far the reward value for the best node we would see after $T$ iterations is from the optimal $f^*$. Or it might be more useful to bound the regret after a given execution time (instead of a number of iterations) in order to compare algorithms that have different computational complexity.

\subsection{Many-armed bandit algorithms}

It is of interest to consider bandit problems in which there are more arms than the possible number of plays, or in which there is an infinity of arms. We refer to this as the ``many-armed'' bandit problem. In this case, we need a model of dependencies between arms in order to get, from one play, information about several arms -- and not only the one that was played. We show how such models can be applied to online global optimisation. In particular, we review the use of Gaussian Processes for modelling arm dependencies.

\subsubsection{Bandits for online global optimisation}

Bandit algorithms have been used to focus exploration in global optimisation. Each point in the space of search is an arm, and rewards are given as we select points where we want to observe the function. Even though the actual objective may not be to minimise the cumulative regret but to minimise the simple regret, we have seen above how a bound on the former can give a bound on the latter. The cumulative regret is also interesting as it forces algorithms not to waste samples. Samples can be costly to acquire in certain applications, as they might involve a physical and expensive action for instance, such as deploying a sensor or taking a measurement at a particular location \citep[see the experiments on sensor networks performed by][]{Srinivas2010}, or they can simply be computationally costly: the less samples, the quicker we can find a maximum.

% Necessity of modelling dependencies when large or infinite number of arms
\paragraph{Modelling dependencies}
The observations may or may not be noisy. In the latter case, the bandit problem is trivial when the search space has less elements than the maximum number of iterations we can perform. But in global optimisation, the search space is usually continuous. In that case, as pointed out by \citet{Wang2008}, if no assumption is made on the smoothness of $f$, the search might be arbitrarily hard. The key idea is to model dependencies between arms, through smoothness assumptions on $f$, so that information can be gained about several arms (if not the whole set of arms) when playing only one arm. Modelling dependencies is also beneficial in problems with finite numbers of arms, as it speeds up the learning. \citet{Pandey2007} have developed an algorithm which exploits cluster structures among arms, applied to a content-matching problem (matching webpages to ads). \citet{Auer2010} use a kernelised version of LinRel, a UCB-type algorithm introduced by \citet{Auer2003} for linear optimisation and further analysed by \citet{Dani2008}, for an image retrieval task with eye-movement feedback. LinRel has a regret in $\tO(\sqrt{T})$, i.e. that grows in $\sqrt{T}$ up to a logarithmic term \footnote{The $\tO$ notation is the one used by \citet{Bubeck2009} and equivalent to $O^*$ used by \citet{Srinivas2010}: $u_n = \tO(v_n)$ iff there exists $\alpha, \beta > 0$ such that $u_n \leq \alpha \log(v_n)^\beta v_n$}.

% Continuous arm spaces and weak-Lipschitz functions
\paragraph{Continuous arm spaces}
Bandit problems in continuous arm spaces have been studied notably by \citet{Kleinberg2008}, \citet{Wang2008} and \citet{Bubeck2009}. To each bandit problem corresponds a mean-reward function $f$ in the space of arms. \citet{Kleinberg2008} consider metric spaces, Lipschitz functions, and derive a regret growth-rate in $O(T^{\frac{d+1}{d+2}})$, which strongly depends on the dimension $d$ of the input space. \citet{Bubeck2009}, however, consider arbitrary topological spaces, weak-Lipschitz functions (i.e. local smoothness assumptions only) and derive a regret in $O(\sqrt{ \exp(O(d)) T})$. The rate of growth is this time independent of the dimension of the input space. Quite interestingly, the algorithm of \citet{Bubeck2009}, HOO, uses BAST on a recursive splitting of the space where each node corresponds to a region of the space and regions are divided in halves, i.e. all non-leaf nodes have two children. BAST is used to select smaller and smaller regions to randomly sample $f$ in. The algorithm developed by \citet{Wang2008}, UCB-AIR, assumes that the probability that an arm chosen uniformly at random is $\epsilon$-optimal scales in $\epsilon^\beta$. Thus, when there are many near-optimal arms and when choosing a certain number of arms uniformly at random, there exists at least one which is very good with high probability. Their regret bound is in $\tO(\sqrt{T})$ when $\beta < 1$ and $f^* < 1$, and in $\tO(T^{\frac{\beta}{1+\beta}})$ otherwise.

\subsubsection{Gaussian Process optimisation}

% Global optimisation and an other smoothness assumption: GP
\paragraph{GP assumption}
In the global optimisation setting, a very popular assumption in the Bayesian community is that $f$ is drawn from a Gaussian Process, due to the flexibility and power of GPs \citep[see][for a review of Bayesian optimisation using GPs]{Brochu2009} and their applicability in practise in engineering problems. GP optimisation is sometimes referred to as ``Kriging'' and response surfaces \citep[see][and references therein]{Grunewalder2010}. GPs are probability distributions over functions, that characterise a belief on the smoothness of functions. The idea, roughly, is that similar inputs are likely to yield similar outputs. The similarity is defined by a kernel/covariance function \footnote{We use the terms `kernel function' and `covariance function' equivalently in the rest of the paper.} between inputs. Parameterising the covariance function translates into a parametrisation of the smoothness assumption. Note that this is a global smoothness assumption which is thus stronger than that of \citet{Bubeck2009}. It is, like the UCB-AIR assumption, a probabilistic assumption too, although a stronger one. \citet{Srinivas2010} claim that the GP assumption is neither too weak nor too strong in practise. One added benefit of this Bayesian framework is the possibility of tuning the parameters of our smoothness assumption (encoded in the kernel) by maximising the likelihood of the observed data, which can be written in closed-form for the commonly used Auto Relevance Determination kernel \citep[see][chap. ̃5]{Rasmussen2006}. In comparison, parameter tuning is critical for HOO to perform well and parameters need to be tuned by hand.

% Acquisition of samples for GP optimisation
\paragraph{Acquisition of samples}
Similarly to bandit problems, function samples are acquired iteratively and it is important to find ways to efficiently focus the exploration of the input space. The acquisition of function samples was often based on heuristics, such as the Expected Improvement and the Most Probable Improvement \citep{Mockus1989} that proved successful in practise \citep{Lizotte2007}. A more principled approach is that of \citet{Osborne2009} which considers a fixed number $T$ of iterations (``finite horizon'' in the bandit terminology) and fully exploits the Bayesian framework to compute at each time step $t$ the expected loss \footnote{In their approach, the loss is defined by the simple regret but one could imagine using the cumulative regret instead.} over all possible $T-t$ remaining allocations as a function of the arm $\xv$ allocated at time $t$. For this, the probability of loss is broken down into the probability of loss given the arms at times $t$ to $T$, times the probability of picking these arms, which can also be broken down recursively. This is similar in spirit to the pioneering work of \citet{Gittins1979} on bandit problems and on ``dynamic allocation indices'' for arms (also known as Gittins index). Here, computing the optimal allocation of $T$ samples has an extremely high computational cost \footnote{In their experiments, the number of iterations was only twice the dimension of the problem.} which is warranted in problems where function samples are very expensive themselves. The simple regret of this procedure was analysed by \citet{Grunewalder2010} in the case where observations are not noisy.

% GP-UCB and regret
\paragraph{UCB heuristic for acquiring samples}
GP approaches have been extended in the bandit setting
\footnote{In practise, rewards are taken in $[0,1]$ in bandit problems, but it is more convenient when dealing with Gaussian Processes to have output spaces centred around $0$ (easier expressions for the posterior mean when the prior mean is the $0$ function). With GPs, we do not assume that the $f$ values are within a known interval. We previously mentioned that an upper bound on $f^*$ could be known, but there is no easy way to encode this knowledge in the prior, which is probably what motivated \citet{Graepel2010} to consider a generalised linear model with a probit link function, in order to learn the Click Through Rates of ads (in $[0,1]$) displayed by web search engines, while maximising the number of clicks (also an exploration vs. exploitation problem).}
, with the Gaussian Process Upper Confidence Bound algorithm (GP-UCB or GPB) presented by \citet{Dorard2009}, for which a theoretical regret bound was given by \citet{Srinivas2010}, based on the rate of decay of the eigenvalues of the kernel matrix on the whole set of arms, if finite, or of the kernel operator: $\tO(\sqrt{T})$ for the linear and Gaussian kernels. This seems to match, up to a logarithmic factor, $T$ times the lower bound on the simple regret given by \citet{Grunewalder2010}, which is a lower bound on the cumulative regret. As the name GP-UCB indicates, the sample acquisition heuristic is based on the optimism in the face of uncertainty principle, where the GP posterior mean and variance are used to define confidence intervals. Better results than with other Bayesian acquisition criteria were obtained on the sensor network applications presented by \citet{Srinivas2010}. There still remains the problem of finding the maximum of the upper confidence function in order to implement this algorithm, but \citet{Brochu2009} showed that global search heuristics are very effective.

%%%
% Purpose and Rationale
%%%

\subsection{A Gaussian Process approach to Tree Search}
In light of this, we consider a GP-based algorithm for searching the potentially very large space of tree paths, with a UCB-type heuristic for choosing arms to play at each iteration. We consider only one bandit problem for the whole tree, where arms are tree paths \footnote{This is similar to \citet[sec. ̃4]{Bubeck2010b} where bandit algorithms for continuous arms spaces are compared to OLOP.}. The kernel used with the GP algorithm is therefore a kernel between paths, and it can be defined by looking at nodes in common between two paths. The GP assumption makes sense for tree search as similar paths will have nodes in common, and we expect that the more nodes in common, the more likely to have similar rewards (this is clearly true for discounted MDPs). Owing to GPs, we can share information gained for `playing' a path with other paths that have nodes in common (which \citealt{Bubeck2010b} also aim at doing as stated in the last part of their Introduction section). Also, we will be able to use the results of \citet{Srinivas2010} to derive problem-independent regret bounds for our algorithm \footnote{The bound will be expressed in terms of the maximum branching factor and depth of the tree, and of the parameters of the kernel in our model, but they won't depend on actual $f$ values.}, once we have studied the decay rate of the eigenvalues of the kernel matrix on the set of all arms (tree paths here), which determines the rate of growth of the cumulative regret in their work.

% Different levels of smoothness with GPs, GP smoothness assumption
\paragraph{Assumptions}
Similarly to BAST, we wish to model different levels of smoothness of the response/reward function on the leaves/paths. For this, we can extend the notion of characteristic length-scale to such functions by considering a Gaussian covariance function in a feature space for paths. Smoothness of the covariance/kernel translates to quick eigenvalue decay rate which can be used to improve the regret bound. As we already said, the parameter(s) of our smoothness assumption can be learnt from training data. Note that the GP smoothness assumption is global, whereas BAST only assumes smoothness for $\eta$-optimal nodes. But in examples such as Go tree search we can expect $f$ to be globally smooth, and for planning in discounted MDPs this is even clearer as $f$ is defined as a sum of intermediate rewards and is thus Lipschitz with respect to a certain metric \citep[see][sec. ̃4]{Bubeck2010b}. As such, $f$ is also made smoother by decreasing the value of the discount factor $\gamma$. Finally, GPs allow to model uncertainty, which results in tight confidence bounds, and can also be taken into account when outputting a sequence of actions at the end of the tree search: instead of taking the best observed action, we might take the one with highest lower confidence bound for a given threshold.

\paragraph{Main results}
We derive regret bounds for our proposed GP-based Tree Search algorithm, run in an anytime fashion (i.e. without knowing the total number of iterations in advance), with tight constants in terms of the parameters of the Tree Search problem. The regret can be bounded with high probability in:
\begin{itemize}
	\item $O(T \sqrt{\log(T)})$ for small values of $T$
	\item $O(\sqrt{T \log(T)})$ for $T \geq B^D$ where $B$ is the maximum branching factor and $D$ the maximum depth of the tree \footnote{$D$ is considered to be fixed but we will see in Section \ref{sec:discussion} that we can extend our analysis to cases where $D$ depends on $T$.}
	\item $O(\log(T) \sqrt{T})$ otherwise.
\end{itemize}
Although the rates are worse for smaller values of $T$, the bounds are tighter because the constants are smaller. For $T \leq B^D$, we have a constant in $\sqrt{\frac{N B}{(B-1) (D+1)}}$ for the linear kernel, and in $\frac{\sqrt{N}}{s}$ for the Gaussian kernel with width $s$: the regret improves when the width increases. Having small constants in terms of the size of the problem is important, since $N = B^D$ is very large in practise and computational budgets do not allow $T$ to go beyond this value.

\subsection{Outline of this paper}

First, we describe the GP-UCB (or GPB) algorithm in greater detail and its application to tree search in Section \ref{sec:algorithm}. In particular, we show how the search for the max of the upper confidence function can be made efficient in the tree case. The theoretical analysis of the algorithm begins in Section \ref{sec:eig} with the analysis of the eigenvalues of the kernel matrix on the whole set of tree paths. It is followed in Section \ref{sec:bound} by the derivation of an upper bound on the cumulative regret of GPB for tree search, that exploits the eigenvalues' decay rate. Finally, in Section \ref{sec:discussion}, we compare GPB to other algorithms, namely BAST for tree search and OLOP for MDP planning, on a theoretical perspective. We also show how a cumulative regret bound can be used to derive other regret bounds. We propose ideas for other Tree Search algorithms based on Gaussian Processes Bandits, before bringing forward our conclusions.

\section{The algorithm}
\label{sec:algorithm}
In this section, we show how Gaussian Processes can be applied to the many-armed bandit problem, we review the theoretical analysis of the GPB algorithm and we describe its application to tree search.

\subsection{Description of the Gaussian Process Bandits algorithm}

We formalise the Gaussian Process assumption on the reward function, before giving the criterion for arm selection in the GPB framework.

\subsubsection{The Gaussian Process assumption}

\paragraph{Definition}
A GP is a probability distribution over functions, and is used here to formalise our assumption on how smooth we believe $f$ to be. It is an extension of multi-variate Gaussians to an infinite number of variables (an $N$-variate Gaussian is actually a distribution over functions defined on spaces of exactly $N$ elements). A GP is characterised by a mean function and a covariance function. The mean is a function on $\Xs$ and the covariance is a function of two variables in this space -- think of the extension of a vector and a matrix to an infinite number of components. When choosing inputs $\xv_a$ and $\xv_b$, the probability density for outputs $y_a$ and $y_b$ is a 2-variate Gaussian with covariance matrix
\begin{equation*}
	\left(\begin{tabular}{cc}
	$\kappa(\xv_a,\xv_a)$ & $\kappa(\xv_a,\xv_b)$\\
	$\kappa(\xv_b,\xv_a)$ & $\kappa(\xv_b,\xv_b)$\\
	\end{tabular}
	\right)
\end{equation*}
This holds when extending to any $n$ inputs. We see here that the role of the similarity measure between arms is taken by the covariance function, and, by specifying how much outputs co-vary, we characterise how likely we think that a set of outputs for a finite set of inputs is, based on the similarities between these inputs, thus expressing a belief on the smoothness of $f$.

% Inference (determining posterior) and noise modelling
\paragraph{Inference and noise modelling}
The reward may be observed with noise which, in the GP framework, is modelled as additive Gaussian white noise. The variance of this noise characterises the variability of the reward when always playing the same arm. In the absence of any extra knowledge on the problem at hand, $f$ is flat \emph{a priori}, so our GP prior mean is the function $\ov$. The GP model allows us, each time we receive a new sample (i.e. an arm-reward pair), to use probabilistic reasoning to update our belief of what $f$ may be -- it has to come relatively close to the sample values (we are only off because of the noise) but at the same time it has to agree with the level of smoothness dictated by the covariance function -- thus creating a \emph{posterior} belief. In addition to creating a `statistical picture' of $f$, encoded in the GP posterior mean, the GP model gives us error bars (the GP posterior variance). In other terms, it gives us confidence intervals for each value $f(\xv)$.

\subsubsection{Basic notations}
% Although notations among the papers referenced in the introduction can be conflicting, we try to use the same notations as Rasmussen for GPs, Srinivas for the GP-UCB algorithm
% Srinivas notations as much as possible (since we are using their results)
%
% \Xs for set of arms (also used by Bubeck), i.e. tree paths, \Ss for states and \As for actions
% balance function: \beta rather than B (used for branching factor)
% confidence threshold: can't use \beta as done by Coquelin, but we use \delta as niranjan does
% \sigma used for posterior std deviation and for noise in niranjan
% we opt for using it for posterior std only, and denote the one for noise by \sn

We consider a space of arms $\Xs$ and a kernel $\kappa$ between elements of $\Xs$. In our model, the reward after playing arm $\xv_t \in \Xs$ is given by $f(\xv_t)+\epsilon_t$, where $\epsilon_t \sim \mathcal{N}(0,\sn^2)$ and $f$ is a function drawn once and for all from a Gaussian Process with zero mean and with covariance function $\kappa$. Arms played up to time $t$ are $\xv_1, \ldots, \xv_t$ with rewards $y_1, \ldots, y_t$. The vector of concatenated reward observations is denoted $\yv_t$. The GP posterior at time $t$ after seeing data $(\xv_1,y_1), \ldots, (\xv_t,y_t)$ has mean $\mu_t(\xv)$ with variance $\s_t^2(\xv)$.

Matrix $C_t$ and vector $\kv_t(\xv)$ are defined as follows:
\begin{eqnarray*}
(C_t)_{i,j} & = & \kappa(\xv_i,\xv_j) + \sn^2 \delta_{i,j}\\
(\kv_t(\xv))_i & = & \kappa(\xv,\xv_i)
\end{eqnarray*}

$\mu_t(\xv)$ and $\s_t^2(\xv)$ are then given by the following equations \citep[see][chap.~2]{Rasmussen2006}:
\begin{eqnarray}
\mu_t(\xv) & = & \kv_t(\xv)^T C_t^{-1} \yv_t \label{eq:posteriorMean}\\
\s_t^2(\xv) & = & \kappa(\xv,\xv) - \kv_t(\xv)^T C_t^{-1} \kv_t(\xv) \label{eq:posteriorVariance}
\end{eqnarray}

\subsubsection{UCB arm selection}
\label{sec:armselection}
The algorithm plays a sequence of arms and aims at optimally balancing exploration and exploitation. For this, we select arms iteratively by maximising an upper confidence function $f_t$:
\begin{equation*}
\xv_{t+1}=\mbox{argmax}_{\xv \in \Xs} \big\{f_t(\xv) = \mu_t(\xv) + \sbet_t \s_t(\xv) \big\}
%\label{eq:argmax}
\end{equation*}
In Section \ref{sec:argmaxft} we show how we can find the argmax of $f_t$ efficiently, in the tree search problem.

\paragraph{Interpretation}
The arm selection problem can be seen as an active learning problem: we want to learn accurately in regions where the function values seem to be high, and do not care much if we make inaccurate predictions elsewhere. The $\sbet_t$ term balances exploration and exploitation: the bigger it gets, the more it favours points with high $\s_t(\xv)$ (exploration), while if $\sbet_t = 0$, the algorithm is greedy. In the original UCB formula, $\sbet_t \sim \sqrt{\log t}$.

\paragraph{Balance between exploration and exploitation}
A choice of $\sbet_t$ corresponds to a choice of an upper confidence bound. \citet{Srinivas2010} give a regret bound with high probability, that relies on the fact that the $f$ values lie between their lower and upper confidence bounds. If $\Xs$ is finite, this happens with probability $1-\delta$ if:
\begin{equation*}
\sbet_t = \sqrt{2 \log(\frac{\abs{\Xs} t^2 \pi^2}{6 \delta})}
%\label{eq:B}
\end{equation*}
However, the constants in their bounds were not optimised, and scaling $\beta_t$ by a constant specific to the problem at hand might be beneficial in practise. In their sensor network application, they tune the scaling parameter by cross validation.

\subsection{Theoretical background}
\label{sec:algorithm:bound}

The GPB algorithm was studied by \citet{Srinivas2010} in the cases of finite and infinite number of arms, under the assumption that the mean-reward function $f$ is drawn from a Gaussian Process with zero mean and given covariance function, and in a more agnostic setting where $f$ has low complexity as measured under the RKHS norm induced by a given kernel. Their work is core to the regret bounds we give in Section \ref{sec:bound}.

\subsubsection{Overview}

\paragraph{Finite case analysis}
When all $f$ values are within their confidence intervals (which, by design of the upper confidence bounds, happens with high probability), a relationship can be given between the regret of the algorithm and its information gain after acquiring $T$ samples (i.e. playing $T$ arms). When everything is Gaussian, the information gain can easily be written in terms of the eigenvalues of the kernel matrix on the training set of arms that have been played so far. The simplest case is for a linear kernel in $d$ dimensions. However, in general there is no simple expression for these eigenvalues since we do not know which arms have been played\footnote{The process of selecting arms is non-deterministic because of the noise introduced in the responses. However, we could maybe determine the probabilities of arms being selected, but such an analysis would be problem-specific (it would depend on $f$ values): we could, as is done in the UCB proof, look at the probability to select an arm given the number of times each arm has been selected so far, and do a recursion... which gives a problem-specific bound in terms of the $\Delta_i$'s.}. Thanks to the result of \citet{Nemhauser1978}, we can use the fact that the information gain is a sub-modular function in order to bound our information gain by the ``greedy information gain'', which can itself be expressed in terms of the eigenvalues of the kernel matrix on the whole set of arms (which is known and fixed), instead of the kernel matrix on the training set. We present this analysis in slightly more detail in Section \ref{sec:algorithm:bound}

\paragraph{Infinite case analysis}
The analysis requires to discretise the input space $\Xs$ (assuming it is a subspace of $\Re^d$), and we need additional regularity assumptions on the covariance function in order to have all $f$ values within their confidence intervals with high probability. The discretisation $\Xs_T$ is finer at each time step $T$, and the information gain is bounded by an expression of the eigenvalues of the kernel matrix on $\Xs_T$. The expected sum of these can be linked to the sum of eigenvalues of the kernel operator spectrum with respect to the uniform distribution over $\Xs$,
% (JST's result)
for which an expression is known for common kernels such as the Gaussian and Matern kernels.
% The latter have been studied by Seeger (?) for the squared-exponential kernel.

\subsubsection{Finite number of arms}

We present two main results of \citet{Srinivas2010} that will be needed in Section \ref{sec:bound}. First, we show that the regret of UCB-type algorithms can be bounded with high probability based on a measure of how quick the function can be learnt in an information theoretic sense: the maximum possible information gain $I^*(T)$ after $T$ iterations (``max infogain''). Intuitively, a small growth rate means that there is not much information left to be gained after some time, hence that we can learn quickly, which should result in small regrets. The max infogain is a problem dependent quantity and its growth is determined by properties of the kernel and of the input space.

Second, we give an expression of the information gain of the ``greedy'' algorithm, that aims to maximise the immediate information gain at each iteration, in terms of the eigenvalues of the kernel matrix on $\Xs$. The max infogain is bounded by a constant times the greedy infogain. We can thus bound the regret in terms of the eigenvalues of the kernel matrix.

\paragraph{Notations}
In the following, $T$ will denote the total number of iterations performed by the algorithm, $R_T$ the cumulative regret after $T$ iterations, $r_t$ the immediate regret at time step $t$, $N = \abs{\Xs}$ the number of arms, $K$ the $N \times N$ kernel matrix on the set of arms, $\av_{i, 1 \leq i \leq N}$ the feature representation of the $i^{th}$ arm, and $\xv$ an element of the feature space (which might not correspond to one of the $\av_i$'s). $I_g$ and $I_u$ will denote respectively the information gain of the greedy algorithm and of the (GP-)UCB algorithm.

% $I^*(T)$ denotes the maximum possible information gain for $A$ of cardinality $T$ (this was denoted by $\gamma_T$ by \citet{Srinivas2010} but we reserve the notation $\gamma$ for the discount factor of the MDP that we will refer to in Section \ref{sec:discussion}).

% \paragraph{Information gain}
% If $A$ is a subset of $\Xs$, getting a sample of outputs $\yv_A$ for the elements in this (training) set reduces our uncertainty about the set of $f$ values for all arms, $\fv_A$. The information gain associated to $A$ is defined as the mutual information between $\yv_A$ and $\fv_A$, i.e. the entropy of $\fv_A$ minus the entropy of $\fv_A$ given $\yv_A$. Everything being Gaussian (Gaussian process and Gaussian noise), the information gain is expressed in terms of the log determinant of $K_A+\sn^2 I$.

\paragraph{Theorem}
 Theorem 1 of \citet{Srinivas2010} uses the fact that GPB always picks the arm with highest UCB value in order to relate the regret to $I_u(T)$:

\begin{equation*}
	\boxed{
R_T \leq \sqrt{\frac{16}{\log(1+\sn^{-2})} \log(\frac{N T^2 \pi^2}{6 \delta}) T I_u(T)} \mbox{ with probability } 1-\delta
	}
\end{equation*}

\paragraph{Greedy infogain}
We define the ``greedy algorithm'' as the algorithm which is allowed to pick linear combinations of arms in $\Xs$, with a vector of weights of norm equal to $1$, in order to maximise the immediate information gain at each time step. An arm in this extended space of linear combinations of the $\av_i$'s is not characterised by an index anymore but by a weight vector. Infogain maximisers are arms that maximise the variance, which is given at arm $\xv = \sum_i v_i \av_i$ by $\vv^T \Sigma_t \vv$ where $\Sigma_t$ is the posterior covariance matrix at time $t$ for the greedy algorithm and $\vv$ is a weight vector of norm 1. Let $\uv_i$ denote the eigenvectors of eigenvalues $\hl_i$ (in decreasing order) of $K$. It can be shown that $\Sigma_t$ and $K$ share same eigenbasis and that the greedy algorithm selects arms such that their weight vectors are among the $t$ first eigenvectors of $K$. The $\Sigma_t$ eigenvalue for the $i^{th}$ eigenvector $\uv_i$ of $K$ is given by:
\begin{equation*}
	\hhl_{i,t} = \frac{\hl_i}{1 + \sn^{-2} m_{i,t} \hl_i}
\end{equation*}
where $m_{i,t}$ denotes the number of times $\uv_i$ has been selected up to time $t$ (we say that a weight vector is selected when the corresponding arm is selected).

Consequently, $\uv_j$ is selected for the first time at time $t$ if all eigenvectors of $K$ of indices smaller than $j$ have been selected at least once and:\\
$\forall i < j, \frac{\hl_i}{1 + \sn^{-2} m_{i,t} \hl_i} \leq \hl_{j}$ (this will be useful in Section \ref{sec:mt0}).

An expression of the greedy infogain can be given in terms of the eigenvalues $\hl_{1 \leq t \leq N}$ of $K$:
\begin{equation*}
	I_g(T) = \frac{1}{2} \sum_{t=1}^{\min(T,N)} \log(1+\sn^{-2} m_t \hl_t)
\end{equation*}
where $m_{t}$ denotes the number of times $\uv_t$ has been selected during the $T$ iterations. We see that the rate of decay of the eigenvalues has a direct impact on the rate of growth of $I_g$.

\paragraph{Maximum possible infogain}
An information-theoretic argument for submodular functions \citep{Nemhauser1978} gives a relationship between the infogain $I_u(T)$ of the GP-UCB algorithm at time $T$ and the infogain $I_g(T)$ of the greedy algorithm at time $T$, based on the constant $e=\exp(1)$:
\begin{equation*}
I^*(T) \leq \frac{1}{1-e^{-1}} I_g(T)
\end{equation*}

As a consequence:
\begin{equation}
	\boxed{
	I_u(T) \leq I^*(T) \leq \frac{1}{2 (1-e^{-1})} \sum_{t=1}^{\min(T,N)} \log(1+\sn^{-2} m_t \hl_t)
	}
	\label{ineq:gammaT}
\end{equation}

It might seem that $I^*(T)$ would always be bounded by a constant because $\min(T,N) \leq N$, which would imply a regret growth in $O(\sqrt{T \log(T)}) = \tO(\sqrt{T})$. However, as we will see in Section \ref{sec:discussion}, we may be interested in running the algorithm with a finite horizon $T$ and letting $N$ depend on $T$. Also, we aim to provide tight bounds for the case where $T \leq N$, with improved constants.
% We will focus on the analysis of $I^*(T)$, which can then be plugged into either of the two regret bounds given above, depending on the assumption on $f$.
The growth rate of $I^*(T)$ might become higher, but the tight constants will result in tighter bounds. Finally, we aim to study how these constants are improved for smoother kernels.

\subsection{Application to tree search}

% D is used for depth here (decision space for niranjan, dimension for Bubeck)
% dim is for dimension

Let us consider trees of maximum branching factor $B \geq 2$ and depth $D \geq 1$. As announced in the introduction, our Gaussian Processes Tree Search algorithm (GPTS) considers tree paths as arms of a bandit problem. The number of arms is $N=B^D$ (number of leaves or number of tree paths). Therefore, drawing $f$ from a GP is equivalent to drawing an N-dimensional vector of $f$ values $\fv$ from a multi-variate Gaussian.

\subsubsection{Feature space}
A path $\xv$ is given by a sequence of nodes : $\xv = x_1, \ldots, x_{D+1}$ where $x_1$ is always the root node and has depth $0$. We consider the feature space indexed by all the nodes $x$ of the tree and defined by
\[
\phi_n(\xv)  =  \left\{\begin{array}{ll}
      1;& \mbox{if } \exists 1 \leq i \leq d, x=x_i\\
0;& \mbox{otherwise.}\end{array}\right.
\]
The dimension of this space is equal to the number of nodes in the tree $N_n = \frac{B^{D+1}-1}{B-1}$.

\paragraph{Linear and Gaussian kernels}
The linear kernel in this space simply counts the number of nodes in common between two paths: intuitively, the more nodes in common, the closer the rewards of these nodes should be. We could model different levels of smoothness of $f$ by considering a Gaussian kernel in this feature space and adapting the width parameter.

\paragraph{More kernels}
More generally, we could consider kernel functions characterised by a set of $\chi_{0 \leq d \leq D}$ decreasing values in $[0,1]$ where $\chi_d$ is the value of the kernel product between two paths that have $d$ nodes not in common. Once the $\chi_{0 \leq d \leq D}$ are chosen, we can give an explicit feature representation for this kernel, based on the original feature space: we only change the components by taking $\sqrt{\chi_{D-i} - \chi_{D-i+1}}$ instead of $1$ if a node at depth $i > 0$ is in the path, and we take $\sqrt{\chi_D}$ at depth $0$ (root). Thus, consider 2 paths that differ on $d$ nodes: the first $1+D-d$ nodes only will be in common, hence the inner product of their feature vectors will be $\chi_D + \sum_{i=1}^{D-d} (\sqrt{\chi_{D-i}-\chi_{D-i+1}})^2 = \chi_d$, which is equal to the kernel product between the 2 paths, by definition of $\chi_d$. Note that the kernel is normalised by imposing $\chi_0 = 1$, which will be required in Section \ref{sec:recursive-block-matrix}.
% Note that the components of the feature vectors stay between $0$ and $1$ if the $\chi$ values are in $[0,1]$

\subsubsection{Maximisation of $f_t$ in the tree}
\label{sec:argmaxft}

The difficulty in implementing the GPB algorithm is to find the maximum of the upper confidence function when the computational cost of an exhaustive search is prohibitive due to a large number of arms -- as for most tree search applications.
%Let us consider the kernel which counts how many nodes two tree paths have in common. $\kappa(\xv,\xv') \leq d$ for all $\xv$ and $\xv'$, and $\kappa(\xv,\xv)=d$ for all $\xv$ ($d$ is the depth of the tree).
At time $t$ we look for the path $\xv$ which maximises $f_t(\xv)$. Here, we can benefit from the tree structure in order to perform this search in $O(t)$ only. We first define some terminology and then prove this result.

\paragraph{Terminology}
A node $x$ is said to be explored if there exists $\xv_{i, i \leq t}$ in the training data such that $\xv_i$ contains $x$, and it is said to be unexplored otherwise. A sub-tree is defined here to be a set of nodes that have same parent (called the root of the sub-tree), together with their descendants. A sub-tree is unexplored if no path in the training data goes through this sub-tree. A maximum unexplored sub-tree is a sub-tree such that its root belongs to an $\xv_i$ in the training data.

\paragraph{Proof and procedure}
$f_t(\xv)$ can be expressed as a function of $\kv = \kv_t(\xv)$ instead of a function of $\xv$ (see Equations \ref{eq:posteriorMean} and \ref{eq:posteriorVariance}) and we argue that all paths that go through a given unexplored sub-tree $S$ will have same $\kv$ value, hence same $f_t$ value. Let $\xv=x_1 \ldots x_l \ldots x_{D+1}$ be such a path, where $l \geq 1$ is defined such that node $x_l$ has been explored but not $x_j$ for $j > l$. All $\xv$'s that go through $S$ have the same first nodes $x_1, \ldots , x_l$, and the other nodes do not matter in kernel computations since they haven't been visited.
%We have by definition: $\kappa(\xv,\xv_i)=\ov$ for all $\xv_i$ which does not contain $x_1$ and $\kappa(\xv,\xv_i)=\max \{ j | x_j \in \xv_i \} \leq l$ otherwise. Hence $\kv_t(\xv)$ does not depend on $\xv \in S$.

Consequently we just need to evaluate $f_t(\xv)$ on one randomly chosen path that goes through the unexplored sub-tree $S$, all other such paths having the same value for $f_t(\xv)$. We represent maximum unexplored sub-trees by ``dummy nodes'' and, similarly to leaf nodes, we compute and store $f_t$ values for dummy nodes. The number of dummy nodes in memory is $1$ per visited node with unexplored siblings: it is the sub-tree containing the unexplored siblings and their descendants. There are at most $D+1$ such nodes per path in the training data, and there are $t$ paths in the training data, hence the number of dummy nodes is less than or equal to $(D+1) t$.

This would mean that the number of nodes (leaf or dummy) to examine in order to find the maximiser of $f_t$ would be in $O(t)$. The search can be made more efficient than examining all these nodes one by one: we assign upper confidence values recursively to all other nodes (non-leaf and non-dummy) by taking the maximum of the upper confidence values of their children. The maximiser of $f_t$ can thus be found by starting from the root, selecting the node with highest upper confidence value, and so on until a leaf or a dummy node is reached. This method of selecting a path is the same as that of UCT and has a cost of $O(B D)$ only. After playing an arm, we would need to update the upper confidence values of all leaf nodes and dummy nodes (in $O(t)$), and with this method we would also need to update the upper confidence values of these nodes' ancestors, adding an extra cost in $O(t)$.

\paragraph{Pseudo-code}
A pseuco-code that implements the search for the argmax of $f_t$ in $O(t)$ is given in Algorithm \ref{alg:gpts}. We sometimes talk about kernel products between leaves and reward on leaves because paths can be identified by their leaf nodes. Note that with this algorithm, we might choose the same leaf node more than once unless $\sn = 0$.

\begin{algorithm} 
\caption{GPB for Tree Search} 
\label{alg:gpts} 
\begin{algorithmic}
	\STATE \COMMENT{Initialisation}
	\STATE create root and dummy child $d_0$
	\STATE $\Ss = \{ d_0 \}$ \COMMENT{set of arms that can be selected}
	\STATE $t = 0$ \COMMENT{number of iterations}
	\STATE $X_t = []$ \COMMENT{leaf-nodes in training set}
	\STATE $\yv_t = []$ \COMMENT{rewards in training set}
	\STATE $K_t = []$ \COMMENT{kernel matrix on the elements of the training set}
	\STATE $C_t = []$ \COMMENT{inverse covariance matrix}
	
	\STATE \COMMENT{Iterations}
	\REPEAT
		\STATE \COMMENT{Choose a path}
		\IF{$t == 0$}
			\STATE $x = d_0$
		\ELSE
			\STATE choose $x$ in $\Ss$ that has highest upper confidence value
		\ENDIF
		\IF{$x$ is a dummy node}
			\STATE \COMMENT{Random walk}
			\STATE create sibling $x'$ of $x$
			\IF{all siblings of $x$ have been created}
				\STATE delete $x$ from the tree and remove from $\Ss$
			\ENDIF
			\STATE $x = x'$
			\WHILE{depth of $x$ is strictly smaller than $D$}
				\STATE create $x'$ child of $x$ and $d$ dummy child of $x$
				\STATE add $d$ to $\Ss$
				\STATE $x = x'$
			\ENDWHILE
			\STATE add $x$ to $\Ss$ \COMMENT{$x$ is the chosen leaf}
		\ENDIF
		
		\STATE \COMMENT{Get reward and add to training set}
		\STATE compute the vector of kernel products $\kv$ between $x$ and the elements of $X_t$
		\STATE append $x$ to $X_t$
		\STATE append $\reward(x)$ to $\yv_t$
		\STATE $K_t =
		\left(\begin{tabular}{cc}
		$K_t$ & $\kv$\\
		$\kv^T$ & $\kappa(x,x)$\\
		\end{tabular}
		\right)$
		\STATE $C_t = (K_t + \sn^2 I_{t+1})^{-1}$
		\FORALL{node $n$ in $\Ss$}
			\STATE compute the vector of kernel products $\kv$ between $n$ and the elements of $X_t$
			\STATE compute the upper confidence of $n$ based on $\kv, C_t, \yv_t$ and Equations \ref{eq:posteriorMean} and \ref{eq:posteriorVariance}
		\ENDFOR
		\STATE $t = t+1$
	\UNTIL{stopping criterion is met}
	
	\STATE \COMMENT{Define output}
	\STATE look for $x$ in $X_t$ that had highest reward value and output the corresponding path

\end{algorithmic}
\end{algorithm}

\section{Kernel matrix eigenvalues}
\label{sec:eig}
For our analysis, we `expand' the tree by creating extra nodes so that all branches have the same branching factor $B$. This construction is purely theoretical as the algorithm doesn't need a representation of the whole tree, nor the expanded tree, in order to run.
% We consider the feature space on tree paths where each dimension corresponds to a tree node and coordinates are binary: $0$ if the node doesn't belong to the path, $1$ otherwise. We consider a normalised linear kernel and a Gaussian kernel in this feature space.

\subsection{Recursive block representation of the kernel matrix}
\label{sec:recursive-block-matrix}

	We write $K_{B,D}$ the kernel matrix on all paths through an expanded tree with branching factor $B$ and depth $D$, and $J_i$ the matrix of ones of dimension $i \times i$. $B$ and $D$ completely characterise the tree (here, nodes don't have labels) so $K_{B,D}$ is expressed only in terms of $B$ and $D$. It can be expressed in block matrix form with $K_{B,D-1}$ and $J_{B^{D-1}}$ blocks:
	\begin{equation}
		K_{B,1} = (\chi_0-\chi_1) I + \chi_1 J_B
		\label{eq:Kinit}
	\end{equation}
	and
	\begin{equation*}
		K_{B,D} = \left(\begin{tabular}{cccc}
		$K_{B,D-1}$ & $\chi_D J_{B^{D-1}}$ & ... & $\chi_D J_{B^{D-1}}$\\
		$\chi_D J_{B^{D-1}}$ & $\ddots$ & $\ddots$ & $\vdots$\\
		$\vdots$ & $\ddots$ & $\ddots$ & $\chi_D J_{B^{D-1}}$\\
		$\chi_D J_{B^{D-1}}$ & ... & $\chi_D J_{B^{D-1}}$ & $K_{B,D-1}$\\
		\end{tabular}
		\right)
	\end{equation*}
	where $\chi_d$ is the value of the kernel product between any two paths that have $d$ nodes not in common.

	To see this, one must think of the $(B,D)$-tree as a root pointing to $B$ $(B,D-1)$-trees. On the 1st diagonal block of $K_{B,D}$ is the kernel matrix for the paths that go through the first $(B,D-1)$-tree. Because the kernel function is normalised, this stays the same when we prepend the same nodes (here the new root) to all paths, so it is $K_{B,D-1}$. Similarly, on the other diagonal blocks we have $K_{B,D-1}$. In order to complete the block matrix representation of $K_{B,D}$ we just need to know that any two paths that go through different $(B,D-1)$-trees only have the root in common, and we use the definition of $\chi_D$.

	Let us denote by $\BI^{(n)}(M)$ and $\BJ^{(n)}(M)$ the matrices of $n$ blocks by $n$ blocks:
	\begin{equation*}
		\BI^{(n)}(M) = \left(\begin{tabular}{cccc}
		$M$ & $0$ & ... & $0$\\
		$0$ & $\ddots$ & $\ddots$ & $\vdots$\\
		$\vdots$ & $\ddots$ & $\ddots$ & $0$\\
		$0$ & ... & $0$ & $M$\\
		\end{tabular}
		\right)
	\end{equation*}
	\begin{equation*}
		\BJ^{(n)}(M) = \left(\begin{tabular}{ccc}
		$M$ & ... & $M$\\
		$\vdots$ & $\ddots$ & $\vdots$\\
		$M$ & ... & $M$\\
		\end{tabular}
		\right)
	\end{equation*}
	We can then write:
	\begin{equation}
		K_{B,D} = \chi_D \BJ^{(B)}(J_{B^{D-1}}) - \chi_D \BI^{(B)}(J_{B^{D-1}}) + \BI^{(B)}(K_{B,D-1})
		\label{eq:Krecurs}
	\end{equation}

\subsection{Eigenvalues}

	For simplicity in the derivations, we consider here the distinct eigenvalues $\bl_{i}$ in increasing order, with multiplicities $\nu_i$. We will later need to ``convert'' these to the $\hl_{1 \leq t \leq N}$ notations used by \citet{Srinivas2010} in order to use their results.

	% We know the eigenvectors and eigenvalues of $J_B$ and, from the $m$ eigenvectors and eigenvalues of $M$, we can deduce those of $\BI_n(M) + l \BJ_n(J_{m})$. We can thus prove with a recursion on $D$ that the eigenvalues of $K_{B,D}$ are:\\
	% $\bl_1 = \chi_0-\chi_1$ with multiplicity $\nu_1 = (B-1) B^{D-1}$\\
	% $\bl_{i+1} = \bl_{i} + B^i (\chi_{i}-\chi_{i+1})$ with multiplicity $\nu_{i+1} = m_{i}/B$\\
	% $\bl_{D+1} = \bl_{D} + B^D \chi_{D}$ with multiplicity $\nu_{D+1} = 1$\\

	We show by recursion that, for all $D \geq 1$, $K_{B,D}$ has $D+1$ distinct eigenvalues $\bl^{(D)}_i$ with multiplicities $\nu^{(D)}_i$:
	\begin{eqnarray}
	\forall i \in [1, D], \bl^{(D)}_i & = & \sum_{j=0}^{i-1} B^j (\chi_j-\chi_{j+1}) \mbox{ and } \nu^{(D)}_i = (B-1) B^{D-i}\\
	\bl^{(D)}_{D+1} & = & \sum_{j=0}^{D-1} B^j (\chi_j-\chi_{j+1}) + B^D \chi_{D} \mbox{ and } \nu^{(D)}_{D+1} = 1
	\end{eqnarray}
	We also show that $J_{B^D}$ and $K_{B,D}$ share same eigenbasis, and the eigenvector $K_{B,D}$ with highest eigenvalue is the vector of ones $\Iv_{B^{D}}$, which is also the eigenvector of $J_{B^D}$ with highest eigenvalue.

	\subsubsection{Proof}
		\paragraph{Preliminary result: eigenanalysis of $J_B$ and $\BJ^{(B)}$}
		$J_B$ has two eigenvalues: $0$ with multiplicity $B-1$ and $B$ with multiplicity $1$. We denote by $\jv_1 ... \jv_{B}$ the eigenvectors of $J_B$, in decreasing order of corresponding eigenvalue. $\jv_1$ is the vector of ones. The coordinates of $\jv_i$ are notated $j_{i,1} ... j_{i,B}$. For all $i$ from $1$ to $B$ we define $\BU_i^{(B)}(.)$ as a concatenation of $B$ vectors:
		\begin{equation*}
			\BU_i^{(B)}(\vv) = \left(\begin{tabular}{c}
			$j_{i,1} \vv$\\
			$\vdots$\\
			$j_{i,B} \vv$\\
			\end{tabular}
			\right)
		\end{equation*}
		For all $i \geq 2$, $\sum_l j_{i,l} = 0$ by definition of $\jv_i$. For all $n$-dimensional vector $\vv$ and $n \times n$ matrix $M$:
		\begin{eqnarray*}
			\BJ^{(B)}(M) \BU_i^{(B)}(\vv) & = & \left(\begin{tabular}{c}
			$(\sum_k M_{1,k} j_{i,1} v_k) + ... + (\sum_k M_{1,k} j_{i,B} v_k)$\\
			$\vdots$\\
			$(\sum_k M_{n,k} j_{i,1} v_k) + ... + (\sum_k M_{n,k} j_{i,B} v_k)$\\
			\end{tabular}
			\right)\\
			& = & \left(\begin{tabular}{c}
			$(\sum_k M_{1,k} v_k) (\sum_l j_{i,l})$\\
			$\vdots$\\
			$(\sum_k M_{n,k} v_k) (\sum_l j_{i,l})$\\
			\end{tabular}
			\right)\\
			& = & \ov
		\end{eqnarray*}
		Hence $\BU_i^{(B)}(\vv)$ is an eigenvector of $\BJ^{(B)}(M)$ with eigenvalue equal to $0$.\\
		
		\paragraph{Recursion}
		We propose eigenvectors of $K_{B,D}$, use Equation \ref{eq:Krecurs} and determine the value of each term of the sum multiplied by the proposed eigenvectors, in order to get an expression for the eigenvalues.
		\begin{itemize}
			\item For $D=1$. From Equation \ref{eq:Kinit}, $\jv_1 ... \jv_{B-1}$ are also eigenvectors of $K_{B,1}$ with eigenvalue $\bl^{(1)}_1 = \chi_0 - \chi_1$, hence $\bl^{(1)}_1$ has multiplicity $\nu^{(1)}_1 = B-1$ as expected. $\jv_B$ is also an eigenvector of $K_{B,1}$ with eigenvalue $\bl^{(1)}_2 = B \chi_1 + \chi_0 - \chi_1$, and $\nu^{(1)}_2 = 1$.\\
			
			\item Let us assume the result is true for a given depth $D-1$. 
			\begin{itemize}
				
				\item The largest eigenvalue of $K_{B,D-1}$ is
				$$\bl^{(D-1)}_D = B^{D-1} \chi_{D-1} + \sum_{j=0}^{D-2} B^j (\chi_j-\chi_{j+1})$$
				with multiplicity $1$. Let us apply $\BU_B^{(B)}$ to the corresponding eigenvector $\Iv_{B^{D-1}}$, and multiply it to the expression of $K_{B,D}$ given in Equation \ref{eq:Krecurs}.\\
				\begin{itemize}
					\item $\BU_B^{(B)}(\Iv_{B^{D-1}}) = \Iv_{B^{D}}$ and $\BJ^{(B)}(J_{B^{D-1}})$ is a matrix of ones in $B^D$ dimensions, hence:
					$$\BJ^{(B)}(J_{B^{D-1}}) \BU_B^{(B)}(\Iv_{B^{D-1}}) = B^D \BU_B^{(B)}(\Iv_{B^{D-1}})$$
					
					\item $\Iv_{B^{D-1}}$ is also the highest eigenvector of $J_{B^{D-1}}$, with eigenvalue $B^{D-1}$, hence:
					$$\BI^{(B)}(J_{B^{D-1}}) \BU_B^{(B)}(\Iv_{B^{D-1}}) = B^{D-1} \BU_B^{(B)}(\Iv_{B^{D-1}})$$
					
					\item By definition of $\Iv_{B^{D-1}}$ and $\bl^{(D-1)}_D$:
					$$\BI^{(B)}(K_{B,D-1}) \BU_B^{(B)}(\Iv_{B^{D-1}}) = \bl^{(D-1)}_D \BU_B^{(B)}(\Iv_{B^{D-1}})$$
				\end{itemize}
				As a consequence, $\BU_B^{(B)}(\Iv_{B^{D-1}}) = \Iv_{B^{D}}$ is the eigenvector of $K_{B,D}$ with highest eigenvalue (this will be confirmed later), equal to $\bl^{(D)}_{D+1} = B^D \chi_{D} + \sum_{j=0}^{D-1} B^j (\chi_j-\chi_{j+1})$.\\
		
				\item Let us apply $\BU_k^{(B)}$ to $\Iv_{B^{D-1}}$ for all $k$ from $1$ to $B-1$.\\
				\begin{itemize}
					\item Owing to the preliminary result, we have:
					$$\BJ^{(B)}(J_{B^{D-1}}) \BU_k^{(B)}(\Iv_{B^{D-1}}) = \ov$$
					\item Since $\Iv_{B^{D-1}}$ is the eigenvector of $J_{B^{D-1}}$ with eigenvalue $B^{D-1}$:
					$$\BI^{(B)}(J_{B^{D-1}}) \BU_k^{(B)}(\Iv_{B^{D-1}}) = B^{D-1} \BU_k^{(B)}(\Iv_{B^{D-1}})$$
					\item Since $\Iv_{B^{D-1}}$ is the eigenvector of $K_{B,D-1}$ with highest eigenvalue:
					$$\BI^{(B)}(K_{B,D-1}) \BU_k^{(B)}(\Iv_{B^{D-1}}) = \bl^{(D-1)}_{D} \BU_k^{(B)}(\Iv_{B^{D-1}})$$ for the same reasons as previously.\\
				\end{itemize}
				As a consequence, $K_{B,D} \BU_k^{(B)}(\Iv_{B^{D-1}}) = (-\chi_D B^{D-1} + \bl^{(D)}_{D+1}) \BU_k^{(B)}(\Iv_{B^{D-1}})$ and we have found $B-1$ eigenvectors of $K_{B,D}$ with eigenvalue equal to $\bl^{(D)}_D = \sum_{j=0}^{D-1} B^j (\chi_j-\chi_{j+1})$. These vectors are also eigenvectors of $J_{B^D}$ with eigenvalue $0$, which comes from the preliminary result and the fact that $J_{B^D} = \BJ^{(B)}(J_{B^{D-1}})$.\\
		
				\item For $i$ from $1$ to $D-1$, let us apply $\BU_k^{(B)}$, for all $k$ from $1$ to $B$, to all $(B-1) B^{D-1-i}$ eigenvectors $\vv$ of $K_{B,D-1}$ with eigenvalue equal to $\bl^{(D-1)}_i$. By definition of $\vv$:
				$$\BI^{(B)}(K_{B,D-1}) \BU_k^{(B)}(\vv) = \bl^{(D-1)}_i \BU_k^{(B)}(\vv)$$
				$\vv$ being also an eigenvector of $J_{B^{D-1}}$ with eigenvalue $0$:
				% $$\forall k \mbox{ s.t. } 1 \leq k \leq B, K_{B,D} \BU_i^{(k)}(\vv) = l \BU_i^{(k)}(\vv)$$
				\begin{eqnarray*}
					\BJ^{(B)}(J_{B^{D-1}}) \BU_k^{(B)}(\vv) & = & \ov\\
					\BI^{(B)}(J_{B^{D-1}}) \BU_k^{(B)}(\vv) & = & \ov
				\end{eqnarray*}
				As a consequence, eigenvalues stay unchanged but their multiplicities are all multiplied by $B$ (because $k$ goes from $1$ to $B$ and we have identified $B$ times as many eigenvectors) which gives $\nu^{(D)}_i = (B-1) B^{D-i}$. Again, the preliminary result allows us to show that the $\BU_k^{(B)}(\vv)$ are also eigenvectors of $J_{B^D}$ with eigenvalue $0$.\\
		
				\item The total number of multiplicities for all found eigenvalues is equal to $(\sum_i^{D-1} \nu^{(D)}_i) + B-1 + 1 = B^D$ so we have identified all the eigenvectors.
		
			\end{itemize}
	
		\end{itemize}

	\subsubsection{Re-ordering of the kernel matrix eigenvalues}
	
		In order to match the notations of \citet{Srinivas2010}, we re-write the eigenvalues as a sequence $\hl_1 \geq \hl_2 \geq ... \geq \hl_N$. We first need to reverse the order of the eigenvalues and thus consider the sequence of $\bl_{D-i}$'s. We obtain the lambda hats by repeating the lambda bars as many times as their multiplicities. $\hl_t = \bl_{D-i}$ with $i$ such that $B^i < t \leq B^{i+1}$. For $1 < t \leq N$, $\log(t)=i\log(B)+r$ with $0<r\leq\log(B)$ hence $B^i < t \leq B^{i+1}$. $i=\frac{\log(t)-r}{\log(B)}$ from which we have:
		\begin{equation}
		\forall t \in [1,N], \exists i \in [-1,D-1], \hl_t = \bl_{D-i} \mbox{ with } \log_B(t)-1 \leq i < \log_B(t)
		\label{ineq:t2i}
		\end{equation}

\subsection{Linear kernel}

	The linear kernel is an inner product in the feature space, which amounts to counting how many nodes in common two paths have. It takes values from $1$ to $D+1$. The normalised linear kernel divides these values by $D+1$. If two paths of depth $D$ differ on $d$ nodes, they have $D+1-d$ nodes in common:
	\begin{equation*}
	\chi_d = \frac{D+1-d}{D+1}
	\end{equation*}

	For all $j$, $\chi_j - \chi_{j+1} = 1/(D+1)$, hence $\bl_i = \frac{1}{D+1} \sum_{j=0}^{i-1} B^j = \frac{B^i-1}{(B-1)(D+1)}$ for $i < D+1$ We use Inequality \ref{ineq:t2i} to get a lower and an upper bound on $\hl_t$ for $t > 1$:
	\begin{eqnarray*}
		& \bl_{D-i} & = \frac{N B^{-i} - 1}{(B-1)(D+1)}\\
		\frac{N B^{-\log_B(t)} - 1}{(B-1)(D+1)} \leq & \hl_t & \leq \frac{N B^{1-\log_B(t)} - 1}{(B-1)(D+1)}
	\end{eqnarray*}
	\begin{equation*}
		\boxed{
		\forall t > 1, \frac{N-t}{(B-1) (D+1) t} \leq \hl_t \leq \frac{N B - t}{(B-1) (D+1) t}
		}
	\end{equation*}
	
	The bounds for $\hl_1$ are obtained by adding $B^D \chi_{D+1}$ to the bounds above. Indeed:
	\begin{eqnarray*}
		\hl_1 & = & \bl_{D+1}\\
		& = & \sum_{j=0}^{D-1} B^j (\chi_j-\chi_{j+1}) + B^D \chi_{D}\\
		& = & \sum_{j=0}^{D} B^j (\chi_j-\chi_{j+1}) + B^D \chi_{D+1}
	\end{eqnarray*}
	We thus see that the expression for $\bl_{D+1}$ only differs from the expressions for other $\bl_i$'s by an added $B^{i-1} \chi_{i}$ term.

\subsection{Gaussian kernel}

We give an expression for $\bl_i$ for this kernel, before giving bounds on $\hl_t$ and studying the influence of the kernel width $s$ on these bounds.

	\subsubsection{Value of $\chi_d$ and $\bl_i$}

		The squared Euclidian distance in the paths feature space is twice the number of nodes $d$ where they differ: path 1 contains nodes indexed by $i_1 ... i_d$ that path 2 doesn't contain, and path 2 contains nodes indexed by $j_1 ... j_d$ that path 1 doesn't contain, so the $i_1 ... i_d$ and $j_1 ... j_d$ components of the feature vectors differ. The components of the difference of the feature vectors will be $0$ except at the $d$ $i$-indices and at the $d$ $j$-indices where they will be $1$ or $-1$. Summing the squares gives $2 d$.

		Consequently, the Gaussian kernel is an exponential on minus the number of nodes where paths differ (from $0$ to $D$):
		\begin{equation*}
		\chi_d = \exp(-\frac{d}{s^2})
		\end{equation*}
		% $\chi_d - \chi_{d+1}$ decreases with $d, d > \s$.

		For all $j$, $\chi_j - \chi_{j+1} = (1-\exp(-\frac{1}{s^2})) \exp(-\frac{j}{s^2})$, hence for all $i < D+1$,
		\begin{eqnarray*}
		\bl_i & = & (1-\frac{\qs}{B}) \sum_{j=0}^{i-1} \qs^j\\
		& = & \Cs (\qs^i-1)
		\end{eqnarray*}
		where
		\begin{eqnarray*}
		\qs & = & B \exp(-\frac{1}{s^2}))\\
		\Cs & = & \frac{1-\frac{\qs}{B}}{\qs-1}
		\end{eqnarray*}
		By definition, $\qs < B$. Let us focus on the case where $1 < \qs$ so that $\Cs$ is always positive, which is equivalent to:
		\begin{equation*}
			s  > \frac{1}{\sqrt{\log(B)}}
		\end{equation*}

	\subsubsection{Bounds on $\hl_t$}

		Once again, Inequality \ref{ineq:t2i} gives us a lower and an upper bound on $\hl_t$:
		\begin{equation*}
			\Cs (q^D q^{-\log_B(t)} -1) \leq \hl_t \leq \Cs (q^D q^{-\log_B(t)} q -1)
		\end{equation*}

		As we will see in the next section, in Inequality \ref{ineq:gammaT}, we are only interested in $t$ indices that are smaller than $N$. As for the linear kernel, we can bound $\hl_t$ by expressions in $1/t$. Indeed:
		\begin{eqnarray*}
		& q^{-\log_B(t)} & = t^{-\log_B(q)}\\
		& q^{-\log_B(t)} & = t^{-1+\frac{1}{s^2 \log(B)}}\\
		\frac{1}{t} \leq & q^{-\log_B(t)} & \leq \frac{1}{t} \exp(\frac{D}{s^2}) \mbox{ since } t \leq B^D\\
		\frac{1}{t} \leq & q^{-\log_B(t)} & \leq \frac{N}{t} q^{-D}
		\end{eqnarray*}

		Which thus gives:
		\begin{equation*}
			\boxed{
			\forall t > 1, \frac{\Cs (N \exp(-\frac{D}{s^2}) - t)}{t} \leq \hl_t \leq \frac{\Cs (N \qs - t)}{t}
			}
		\end{equation*}
		for $s > \frac{1}{\sqrt{\log(B)}}$.

	\subsubsection{Influence of the kernel width}
	
		From the above we have:
		\begin{equation*}
			\hl_t \leq \frac{N \Cs \qs}{t}
		\end{equation*}
		Note that
		\begin{eqnarray*}
		\Cs \qs & = & \frac{(B-\qs) \qs}{B (\qs-1)}\\
		& = & (1+\frac{1}{\qs-1}) (1-\frac{\qs}{B})
		\end{eqnarray*}
		and $\qs$ increases when $s$ increases, hence $\frac{1}{\qs-1}$ decreases and $-\qs$ decreases. As a result, $\Cs \qs$ decreases. Also, since $\qs$ tends to $B$ when $s$ tends to infinity, the limit of $\Cs \qs$ is $0$ when $s$ tends to infinity. The $\hl_t$ upper-bound improves over that of the linear kernel when $s$ is big enough so that 
		$\Cs \qs \leq \frac{B}{(B-1) (D+1)}$.
		
		Now, let us look at the rate at which $\Cs \qs$ tends to zero: when $s$ is bigger than $\frac{1}{\sqrt{\log(\frac{B}{2})}}$, we have:
		\begin{eqnarray*}
		\Cs \qs & \leq & 2 (1-\exp(-\frac{1}{s^2}))\\
		& \leq & 2 (\frac{1}{\s^2} + o(\frac{1}{s^2}))
		\end{eqnarray*}
		Hence:
		\begin{equation*}
			\boxed{
				\Cs \qs = O(\frac{1}{s^2})
			}
		\end{equation*}

		% As a more general remark, for a fixed interval of $\chi$ values, the slower $\chi_d$ decreases as a function of $d$, the slower $\chi_j - \chi_{j+1}$ decreases as a function of $j$, the quicker $\bl_i$ increases and the quicker $\hl_t$ decreases. ???

\section{GP Tree Search bounds}
\label{sec:bound}
Information gain bounds can be turned into high probability regret bounds as seen in Section \ref{sec:algorithm}. In Section \ref{sec:kernel-indep} we give two information gain bounds which are valid for any kernel with $\chi$ values in $[0,1]$. Bound \ref{ineq:bound1} is independent of time and is interesting asymptotically. However, in most interesting tree search problems, $N$ is extremely large (consider $B=200$ as in Go, and $D=10$) and the number of iterations $T$ is smaller than $N$. Bound \ref{ineq:bound3} is log linear in $T$ but only involves constants that are small compared to $N$ (in $O(B D)$), which is interesting when $T<N$.

In Section \ref{sec:kernel-spec} we derive better bounds that take advantage of the decay of the kernel matrix eigenvalues. Bounds \ref{ineq:boundLog} and \ref{ineq:boundTail}, when combined, give a function which is linear up to a time $T_*$, after which it becomes logarithmic in the number of iterations. For $T > N$ the bound becomes independent of time. We show that, in the case of the Gaussian kernel, the constants improve in $O(\frac{1}{s^2})$ when the kernel width $s$ increases. The Gaussian kernel bound can be better than the linear kernel bound for large enough values of $s$.

\subsection{Kernel-independent bound}
\label{sec:kernel-indep}
We run the algorithm until time $T$. We could use the feature representation described in the previous section, in which case the dimension would be bounded by the number of visited nodes which is itself bounded by $D\ T$. Alternatively, we could consider feature representations which are indicator vectors, in which case the dimension would be bounded by $T$ (we have seen $T$ different arms at most). Doing so equates to changing the kernel matrix into the identity matrix, which can only worsen the regret since arms become independent. It can be preferable to switch to the feature representation described in the previous section for large values of $T$, since the number of visited nodes is bounded by $N_n$.

The algorithm runs with a linear kernel in this feature space, or, if we consider the Gaussian kernel, a slightly different feature space with same dimensionality (as seen in the previous section). This corresponds to case 1 of Theorem 5 from \citet{Srinivas2010}. For a linear kernel, $\kappa(\xv,\xv')=\xv^T \xv'$ and the kernel matrix $K_T$ on training set $[\xv_1 ... \xv_T]$ is equal to $X_T^T X_T$. Let us denote by $\Lambda$ the diagonal matrix of eigenvalues $\bar{l}_1 \geq ... \geq \bar{l}_{dim}$ of $X_T X_T^T$. The information gained from a training set $A_T$ can be expressed in terms of $K_T$:
\begin{eqnarray*}
F(A_T) % & = & H(\fv_\Xs) - H(\fv_\Xs | \yv_T)\\
%  & = & H(\fv_T) - H(\fv_T | \yv_T)\\
 & = & H(\yv_T) - H(\yv_T | \fv_T)\\ %\mbox{ by Bayes' law}
 & = & H(N(\ov, K_T + \sn^2 I_T)) - H(N(\fv_T, \sn^2 I_T))\\
% & = & 1/2 \log(\abs{K_T + \sn^2 I_T}) - 1/2 \log(\abs{\sn^2 I_T})\\
% & = & 1/2 \log(\frac{\abs{K_T + \sn^2 I_T}}{\sn^{2T}})\\
 & = & 1/2 \log(\abs{I_T + \sn^{-2} X_T^T X_T})\\
 & = & 1/2 \log(\abs{I + \sn^{-2} X_T X_T^T}) \mbox{ by Sylvester's determinant theorem}\\
 & \leq & 1/2 \log(\abs{I + \sn^{-2} \Lambda}) \mbox{ by Hadamard's inequality}\\
 & \leq & \sum_{i=1}^{dim} 1/2 \log(1+\sn^{-2} \bar{l}_i)\\
% & \leq & \sum_{i=1}^{dim} 1/2 \log(1+\sn^{-2} \bar{l}_1)\\
 & \leq & (dim/2) \log(1+\sn^{-2} dim)
\end{eqnarray*}
This results from the fact that $X_T X_T^T$ is a $dim \times dim$ real symmetric matrix with entries in $[0,1]$, hence every eigenvalue is smaller than $\bar{l}_1$ which itself is smaller than $dim$ \citep[see][]{Zhan2005}. Using $dim \leq T$ and maximising over $A_T$, we get:

\begin{equation}
\label{ineq:bound3}
I^*(T) \leq (T/2) \log(1+ \sn^{-2} T)
\end{equation}

We could have a bound in $O(\log(T))$ instead by bounding the first occurrence of $dim$ by $N_n$, but the constant would be huge ($O(N)$) and would make the result less interesting for $T < N$. We could even have a bound independent of time when bounding both occurrences of $dim$ by $N_n$:

\begin{equation}
\label{ineq:bound1}
I^*(T) \leq (N_n/2) \log(1+ \sn^{-2} N_n)
\end{equation}

These bounds do not depend on the $\chi_d$ values and are therefore kernel-independent.

% \subsection{Identity kernel matrix}
% 
% All eigenvalues are $1$. We get:
% \begin{eqnarray}
% I^*(T) & \leq & \frac{1/2}{1-e^{-1}} (T \log(2 \sn^{-2} T) + T \log(1))
% \end{eqnarray}
% 
% The growth of $I^*(T)$ is $O(T \log(t))$ (same as in the first bound) but the constant here doesn't depend on $D$ nor $B$.

\subsection{Sum of log-eigenvalues bound}
\label{sec:kernel-spec}

We start from Inequality \ref{ineq:gammaT}.
% We split the sum into two parts, such that the $\bl_{t \leq T_*}$ contain most of the mass of all the eigenvalues, and the $\bl_{t > T_*}$ represent a smaller fraction of this mass.
Without further knowledge on the $m_t$'s, we simply lower-bound them by $0$ and upper bound by $T$.

% \begin{eqnarray*}
% \sum_{t=1}^{T_*} \log(1+\sn^2 m_t \hl_t) & \leq & T_* \log(1+\sn^2 r N) \mbox{ where } r = \sum_{t \leq T_*} m_t\\
% \sum_{t=T_*+1}^{T} \log(1+\sn^2 m_t \hl_t) & \leq & \sn^2 (T-r) \sum_{t=T_*+1}^{T} \hl_t
% \end{eqnarray*}

In order to exploit our upper-bound on $\hl_t$, we first bound the sum of $\log(1+\sn^{-2} m_t \hl_t)$ by a sum of $\log(c' \hl_t) = \log(c') + \log(\hl_t)$ so that a sum of $\log(\hl_t)$ appears. This can in turn be bounded owing to a result on the sum of $\log(1/t)$:
\begin{eqnarray}
\sum_{t=2}^{\min(T,N)} \log(1/t) & \leq & \sum_{t=1}^{T} \log(1/t)\\
 & \leq & \log(1/T!)\\
 & \leq & - \log(\Gamma(T+1))\\
 & \leq & - T \log(\frac{T+1}{e})
\end{eqnarray}
using the fact that $\Gamma(x) \geq (\frac{x}{e})^{x-1}$.

Let us consider the Gaussian kernel \footnote{The derivations for the linear kernel are very similar and just involve different constants.}. $\hl_{T}$ being the smallest eigenvalue and $m_t$ being either zero or bigger than $1$, we have either $\log(1+\sn^{-2} m_t \hl_t) = 0$ or $\frac{m_t \hl_t}{\hl_{T}} \geq 1$. From there:

% $\hat{\bl_t} \geq \hl_{T_m} \geq \hl_N = \bl_1$ because $t \leq T_m$.
% $m_t \frac{\hat{\bl_t}}{\bl_1} \geq \frac{\hat{\bl_t}}{\bl_1} \geq 1$ because $m_t \geq 1$

\begin{eqnarray*}
\log(1+\sn^{-2} m_t \hl_t) & \leq & \log((\frac{1}{\hl_T} + \sn^{-2}) m_t \hl_t) \mbox{ for $t$ s.t. } m_t \neq 0\\
& \leq & \log((\frac{1}{\hl_T} + \sn^{-2}) T \hl_t) \mbox{ for all $t$}\\
% & \leq & \log((\frac{1}{\bl_1} + \sn^{-2}) T N C(\s)) - \log(t+1)\\
\sum_{t=1}^{\min(T,N)} \log(1+\sn^{-2} m_t \hl_t) & \leq & \log((\frac{1}{\hl_T} + \sn^{-2}) T N \Cs \qs) T + \sum_{t=2}^{\min(T,N)} \log(\frac{1}{t}) + \log(\hl_1)\\
% & \leq & T \log((\frac{1}{\bl_1} + \sn^{-2}) T N C(\s)) - (T+1) \log(\frac{T+2}{e})\\
% & \leq & \log(\frac{(\frac{1}{\hl_T} + \sn^{-2}) T N \Cs e}{T+1}) T\\
% & \leq & \log((\frac{1}{1-\exp(-1/s^2)} + \sn^{-2}) N C(\s) e) (T+1)\\
% & \leq & \log(\frac{1}{B \exp(-1/s^2)-1} N e + \sn^{-2} N C(\s) e) (T_*+1)
\end{eqnarray*}

\begin{equation}
	\boxed{
	I^*(T) \leq \frac{1}{2(1-e^{-1})} \log((\frac{1}{\hl_T} + \sn^{-2}) N \Cs \qs e) T + D \log(B)
	}
	\label{ineq:boundLog}
\end{equation}

By extracting $\log(\hl_t)$ terms from $I^*(T)$, we take advantage of the $\log$ but we also introduce a $\frac{1}{\hl_{T}}$ term and thus we suffer from smooth kernels for which $\hl_{T}$ will be low.

\subsection{Eigenvalues tail-sum bound} % for $T_m \leq T \leq N$

To remedy this, we split the sum at $T_*$ such that the $\hl_{t > T_*}$ are low and it is acceptable to bound $\log(1+\sn^{-2} m_t \hl_t)$ by $\sn^{-2} m_t \hl_t$. Thus, we consider the tail-sum of the eigenvalues which allows us to exploit quick decay rates for smooth kernels, resulting in small regret bounds.
% The tail sum can be bounded by a linear expression of the eigenvalues, which is acceptable for low eigenvalues, i.e. for $T_*$ big enough. Bounding this sum by a sum of the eigenvalues allows us, unlike the previous bound on the sum to $T_*$, to exploit smooth kernels that have quicker eigenvalues decay rates. % In this form, we already see that the smoother the kernel, the quicker the eigenvalues will decay, and the better the regret bound.

$\hl_t \leq N \hat{L}(t)$ where $\hat{L}$ is a decreasing function, so that we can bound the tail-sum $\sum_{t=T_*+1}^T \hl_t$ by $N$ times a tail integral of $\hat{L}$
%\footnote{Ideally, $T_* < T$ should be such that the integral is small enough to compensate for the very large $N$ factor.}
. The quicker $\hat{L}$ decreases, the lower the integral, hence the lower the information gain bound and the lower the regret. For the Gaussian kernel and $s > \frac{1}{\log(B)}$:

\begin{equation*}
\sum_{t=T_*+1}^{\min(T,N)} \hl_t \leq N \int_{t=T_*}^{\min(T,N)} \hat{L}(t) dt
\end{equation*}

\begin{equation}
	\boxed{
	I^*(T) - I^*(T_*) \leq \frac{N}{2 (1-e^{-1})} \Cs \qs \log(\frac{\min(T,N)}{T_*})
	}
	\label{ineq:boundTail}
\end{equation}

\subsection{Sum to $T'$ where $\forall t > T', m_t=0$}
\label{sec:mt0}

We know that the greedy procedure chooses eigenvectors of $K$ among the $T$ that have highest associated eigenvalue. However, we might have only picked $T'$ eigenvectors and picked several times the same ones ($m_i$ gives the number of times we have picked the $i^{th}$ biggest eigenvector of $K$). We look for the smallest $T'$ such that:
\begin{equation}
\label{prop:p}
\forall T' < t \leq T, m_t=0
\end{equation}
The information gain will then be bounded by $\sum_{t=1}^{T'} \log(1+\sn^{-2} m_t \hl_t)$.

The contrary of Proposition \ref{prop:p} is equivalent to choosing $\uv_{T'+1}$ at least once. This is equivalent to the fact that there exists $t$, first time we select $\uv_{T'+1}$, such that all eigenvalues $\hhl_{i,t}$ of $\Sigma_t$ are smaller than $\hhl_{T'+1,t} = \hl_{T'+1}$. This can be written:
\begin{eqnarray}
	\exists t \leq T, \forall i \leq T', \frac{\hl_i}{1 + \sn^{-2} m_{i,t} \hl_i} & \leq & \hl_{T'+1}\\
	\frac{1}{\hl_{T'+1}} - \frac{1}{\hl_i} & \leq & \sn^{-2} m_{i,t}
	\label{prop:q}
\end{eqnarray}

Therefore, Not Proposition \ref{prop:p} is equivalent to Proposition \ref{prop:q}. Let us assume that the latter is true. We know that each $m_{i,t}$ is smaller than $m_{i,T}$ and that $\sum_{i=1}^{T'} m_{i,T} \leq T$, hence:
\begin{equation}
	\sum_{i=1}^{T'} (\frac{1}{\hl_{T'+1}} - \frac{1}{\hl_i}) \leq \sn^{-2} T
	\label{prop:r}
\end{equation}

Thus, we can find $T'$ such that Proposition \ref{prop:p} is true by lower bounding $\sum_{i=1}^{T'} (\frac{1}{\hl_{T'+1}} - \frac{1}{\hl_i})$ and looking for $T'$ such that this lower bound is equal to $\sn^{-2} T$.

From the $\hl_t$ bounds established in the previous section, we have:
\begin{eqnarray*}
	\sum_{i=1}^{T'} (\frac{1}{\hl_{T'+1}} - \frac{1}{\hl_i}) & \geq & \sum_{i=1}^{T'} \frac{T'+1}{N \Cs \qs} - \frac{i}{\Cs (N \exp(-\frac{D}{s^2}) - i)}\\
	& \geq & \frac{1}{\As} T' (T'+1)\\
	\mbox{where } \As & = & \frac{N \Cs \qs (\exp(-\frac{D}{s^2}) - 1)}{\exp(-\frac{D}{s^2}) - 1 - \frac{\qs}{2}}
\end{eqnarray*}
thus we look for $0 < T' < T$ such that $T'^2 + T' - \sn^{-2} \As T = 0$:
\begin{equation*}
	T' = \frac{-1 + \sqrt{1 + 4 \sn^{-2} \As T}}{2} = O(\sqrt{T})
\end{equation*}
For $T$ big enough, the previous expression is smaller than $T$. Also, since $\As$ tends to $0$ when $s$ tends to infinity, we know that the bigger $s$, the smaller the constant in front of $\sqrt{T}$ in the expression for $T'$, and thus the earlier $T' < T$ is true.

When replacing $T$ by $T'$ in the $\log$ of the info gain bound, we can divide the constant in front of the $\log$ by $2$ while slightly increasing the offset, but we cannot improve the growth rate in $T$.

\subsection{Final bound}

Increasing $T_*$ will decrease the eigenvalues tail-sum bound, but it will increase the sum of log-eigenvalues bound. Conversely, decreasing $T_*$ increases the former bound and decreases the latter. We know we picked the optimal $T_*$ when the cost to put $T_*+1$ in the first sum would be higher than to put it in the tail sum (regardless of the value of $T$):
\begin{eqnarray*}
\log((\frac{1}{\hl_{T_*+1}} & + & \sn^{-2}) N \Cs \qs e) (T_*+1) - N \Cs \qs \log(T_*+1)\\
& \geq & \log((\frac{1}{\hl_{T_*}} + \sn^{-2}) N \Cs \qs e) T_* - N \Cs \qs \log(T_*)
\end{eqnarray*}

$T_*$ doesn't depend on $T$ but on the kernel: the smoother the kernel, the smaller $T_*$. For $T \leq T_*$, we will only be using the sum of log-eigenvalues bound, resulting in a linear information gain bound. For $T > T_*$ we will combine the two bounds. The first bound becomes a constant when replacing $T$ by $T_*$, and the second bound dictates the rate of growth of $I^*(T)$: $\log(T)$ when $T \leq N$ and constant otherwise. Thus, the regret is in $O(T \sqrt{\log(T)})$ or in $O(\log(T) \sqrt{T})$ or in $O(\sqrt{T \log(T)})$. The last two growth rates can be rewritten as $\tO(\sqrt{T})$.

\subsubsection{Influence of $s$}
The regret should improve for smoother kernels, i.e. bigger $s$. Let us check that this is the case with the bounds we gave. As we already noted, $\Cs \qs$ decreases in $O(\frac{1}{s^2})$. Hence, the eigenvalues tail-sum bound is clearly improving for larger values of $s$. Let us now derive a sum of log-eigenvalues bound in terms of $s$:
\begin{eqnarray*}
	\sum_{t=1}^{T} \log(1+\sn^{-2} m_t \hl_t) & \leq & \log((\frac{1}{\hl_T} + \sn^{-2}) N \Cs \qs e) T\\
	& \leq & \log(\frac{N \Cs \qs e T}{\Cs (N \exp(-\frac{D}{s^2}) - T)} + \sn^{-2} N \Cs \qs e) T\\
	& \leq & \log(\frac{B \exp(-\frac{1}{s^2}) e T}{\exp(-\frac{D}{s^2}) - 1} + \sn^{-2} N \Cs \qs e) T\\
	& \leq & \log(\frac{B e T}{\exp(-\frac{D-1}{s^2}) - \exp(\frac{1}{s^2})} + \sn^{-2} N \Cs \qs e) T
\end{eqnarray*}

The first term of the sum inside the log is decreasing as $\exp(-\frac{D-1}{s^2})$ and $- \exp(\frac{1}{s^2})$ are increasing, and the second term of the sum is also decreasing as $\Cs \qs$ is decreasing.

%\section{Experiments}
%\label{sec:experiments}
%\input{experiments}

\section{Discussion}
\label{sec:discussion}
In this section, we discuss the GPTS algorithm and our previous results in relation to other algorithms for tree search and planning in MDPs. We then give a few ideas to extend our work and conclude this paper.

\subsection{Tree Search}
	
	We compare GPTS to the Bandit Algorithm for Smooth Trees algorithm.
			
		\subsubsection{Extension of $f$ to non-leaves}
			\citet{Coquelin2007a} extend the definition of $f$ to all nodes: let us call $\hat{f}$ the function which coincides with $f$ on the leaf nodes and which, on any other node $n$, is equal to the maximum value of $f$ on tree paths that go through $n$.
			% The $\hat{f}$ values can be computed recursively: for all $B$ nodes at depth $D-i$ having same parent, $\hat{f}$ at this parent (depth $D-i-1$) is the maximum of the $\hat{f}$ values on these nodes (i.e. the $f$ values for $i=0$).
			The maximum value of $\hat{f}$ is $\hat{f}^* = f^*$. We can also extend the definition of the suboptimality $\Delta_n = f^* - f(n)$ of a leaf node to any node $n$ at any depth: $\hat{\Delta}_n = \hat{f}^* - \hat{f}(n)$. The BAST smoothness assumption on $\hat{f}$ is that, for any $\eta$-suboptimal node $n$ at depth $d$ (meaning $\hat{\Delta}_n \leq \eta$), there exists $\delta_d > 0$ such that $\hat{f}(n)-\hat{f}(i) \leq \delta_d$ for any child $i$ of $n$. This is only a local regularity assumption as there is no assumption on nodes which are not $\eta$-suboptimal.
			
		\subsubsection{Smoothness of $f$}
			For two given nodes $n_1$ and $n_2$ with same parent, there exist two leaves $l_1$ and $l_2$ (with ancestors $n_1$ and $n_2$, respectively) such that $\hat{f}(n_1)-\hat{f}(n_2) = f(l_1) - f(l_2)$. With the GPB assumption, $(f(l_1),f(l_2))^T$ lies with high probability within an ellipse determined by the kernel between $l_1$ and $l_2$ (equal to the depth of $n_1$ and $n_2$, when considering the linear kernel). One can thus say how close the $\hat{f}$ values of two siblings may be, and thus bound $\hat{f}(n)-\hat{f}(i)$ in terms of the depth of $n$, with high probability, in order to give a rough comparison with the BAST smoothness assumption. Although this bound is only with high probability -- while it would always hold with BAST -- GPB makes an extra assumption on how the $\hat{f}$ values would be distributed.
			
		\subsubsection{Reward variability}
			BAST assumes that the reward at each leaf is always in $[0,1]$ and is given by a probability distribution with mean equal to the $f$ value at that leaf, whereas GPB assumes that the reward distribution is Gaussian with standard deviation $\sn$. However, \citet{Srinivas2010} have also extended their regret analysis to the more general case where the rewards are given by $f(\xv_t) + \epsilon_t$ such that the sequence of noise variables $\epsilon_t$ is an arbitrary martingale difference sequence uniformly bounded by $\sn$. The resulting regret has same growth rate in $T$.
			
		\subsubsection{Regret bounds}
			Theorem 4 of \citet{Coquelin2007a} gives a regret bound when $\delta_d$ decreases exponentially: $\delta_d = \delta \gamma^d$. The bound is written in terms of the parameters of the smoothness assumption (namely $\eta, \delta, \gamma$) and is independent of time. However, this bound is problem-specific as it involves the inverse of the $\Delta_{min}$ quantity, where $\Delta_{min} = \min_i \{\Delta_i = f^*-f(i)\}$. Note that when $f$ has $B^D$ possible inputs, $1/\Delta_{min}$ can easily be of the order of $B^D$.

			While the BAST bound may be interesting asymptotically, the number of iterations $T$ of the tree search algorithm is unlikely to go past $B^D$ for most interesting values of $B$ and $D$.
			% Hence, it may be preferable to have a bound that depends on $T$ but not on $1/\Delta_{min}$, so that the bound is tighter for smaller values of $T$.
			An issue with the $1/\Delta_{min}$ term is that, the smoother $f$, the bigger $1/\Delta_{min}$ and the bigger the regret -- whereas we actually would like to take advantage of the smoothness of $f$ to improve (decrease) the regret. The non-dependency w.r.t. $\Delta_{min}$ usually comes at the price of a stronger dependency on time $T$, as it is the case with UCB \footnote{For UCB, the problem-specific bound is in $O(\log(T))$ while the problem-independent bound is in $O(\sqrt{T})$ \citep[see][]{Bubeck2010a}}.
		
			% Tree growing method
		\subsubsection{Tree growing method}
		
			\paragraph{Iterative-deepening}
			The trees we set to search are usually too big to be represented in memory, which is why we ``grow'' them iteratively by only adding the nodes that are needed for the implementation of our algorithm. One method of growing the tree is by iterative-deepening, used for Go tree search by \citet{Coulom2006}: the current iteration is stopped after creating a new node (or reaching a maximum depth); a reward is obtained as a function of the visited path (not necessarily of depth $D$), or as a function of a randomly completed path of length $D$. The resulting tree is asymmetric and contains paths that have different numbers of nodes. Hopefully this helps to go deeper in the tree in regions where $f$ has high values, and keeps the paths short in the rest of the tree. This saves time and memory by stopping the exploration at a depth smaller than $D$ and not creating nodes that would belong to sub-optimal paths.
		
			\paragraph{Depth-first}
			Because we consider tree paths as arms of a bandit problem in GPTS, we need all paths to have same length, which results in a depth-first tree growing method. Depth-first means that, at each iteration, we add nodes sequentially until reaching a maximum depth $D$, and then we start another iteration of the tree search algorithm from the root. So, unlike GPTS, BAST can be run in either iterative-deepening or depth-first mode. Supposedly the algorithm is more efficient in its iterative-deepening version, but no regret bound was given for this version. % -- \citet{Coquelin2007a} only give a problem-specific bound on the number of times nodes are visited.

\subsection{Open loop planning in MDPs}

	We compare GPTS to the Open Loop Optimistic Planning algorithm. In Tree Search applied to planning in MDPs, the reward is a sum of discounted intermediate rewards. \citet{Bubeck2010b} assume that these intermediate rewards are bounded in $[0,1]$, but it is better to translate them to $[-1,1]$ if we plan to use the GPTS algorithm (so that the prior mean for $f$ can indeed be taken to be $\ov$).
		
		\subsubsection{Choice of a GP model for rewards in discounted MDPs}
		% Is this coherent with intro on MDPs?
		We model our belief on what we expect the intermediate reward functions to be, by considering, at each node $n_\tau$ in the sequence of actions being explored, a set of random variables $F^{(n_\tau)}_1, ... F^{(n_\tau)}_B$ such that the intermediate reward function values for all possible actions from node $n_\tau$ is a realisation of this. We assume that each of these random variables follows a normalised Gaussian distribution, and that they are all independent. We now determine the tree paths kernel function that follows from this assumption. A path is a list of nodes $n_0, n_1, ..., n_D$, where $n_0$ is the root, corresponding to a list of indices $i_1, ... i_{D}$ of actions taken in the environment. Our belief on the function value for this path is represented by $\gamma^0 F^{(n_0)}_{i_1} + ... + \gamma^{D-1} F^{(n_{D-1})}_{i_D}$. If two paths $\xv$ and $\xv'$ have $h$ action indices in common, they can be represented by $i_1, ... i_{h}, i_{h+1}, ... i_{D}$ and $i_1, ... i_{h}, i'_{h+1}, ... i'_{D}$. The kernel product between these two paths is given by:
		\begin{eqnarray*}
			\kappa(\xv,\xv') & = & \cov(\gamma^0 F^{(n_0)}_{i_1} + ... + \gamma^{h-1} F^{(n_{h-1})}_{i_h} + \gamma^{h} F^{(n_h)}_{i_{h+1}} + ... + \gamma^{D-1} F^{(n_{D-1})}_{i_D},\\
			& & \gamma^0 F^{(n_0)}_{i_1} + ... + \gamma^{h-1} F^{(n_{h-1})}_{i_h} + \gamma^{h} F^{(n'_h)}_{i'_{h+1}} + ... + \gamma^{D-1} F^{(n'_{D-1})}_{i'_D})\\
			& = & \sum_{\tau=0}^{h-1} \gamma^l \gamma^l \cov(F^{(n_\tau)}_{i_{\tau+1}}, F^{(n_\tau)}_{i_{\tau+1}})\\
			& = & \frac{1-\gamma^{2h}}{1-\gamma^2}
		\end{eqnarray*}
		where we used the bilinearity of the covariance, the independence of the random variables, and the fact that their variances are always $1$. This characterises our belief on the discounted sum of rewards $f$. Note that the kernel is not normalised: $\kappa(\xv,\xv) = \frac{1-\gamma^{2D}}{1-\gamma^2}$ which grows with $D$. This reflects the fact that the signal variance is higher for deeper trees.
		
		Although OLOP and BAST are very similar in spirit, OLOP exploits the fact that $f$ is globally smooth, owing to the discount factor $\gamma$. Again, the GP smoothness assumption is weaker in the sense that the intermediate rewards are not bounded, but it is stronger since we make an assumption on how they are distributed. However, previous studies on Bayesian optimisation gives us reasons to think that this may be reasonable in practise.

		\subsubsection{Simple regret}
		\citet{Bubeck2010b} consider the simple regret $f^* - f_{best(T)}$ as a more appropriate measure of performance for a planning algorithm, for which they obtain a bound by dividing a cumulative regret bound by $T$. Since the algorithm outputs the best observed path, it might actually be interesting to give a bound on the empirical simple regret, i.e. on $|f^* - y_{best(T)}|$. A relationship between the empirical simple regret and cumulative regret can be given with high probability. First, we define the empirical cumulative regret as $R'_T = T f^* - \sum_{t=1}^T y_t$. \citet{Coquelin2007a} give a relationship between the cumulative regret and the empirical cumulative regret: $|R_T - R'_T| = O(\sqrt{T})$ with high probability. Finally:
		\begin{equation*}
			f^* - y_{best(T)} \leq \frac{1}{T} R'_T
			\leq \frac{1}{T} R_T + O(\frac{1}{\sqrt{T}})
		\end{equation*}
		with high probability.
		
		\subsubsection{Regret bounds}
		The cumulative regret considered by \citeauthor{Bubeck2010b} is measured as a function of the number of calls $n$ to the generative model, which is equal to $D\ T$ for us. Their immediate regret for a given policy is defined as the difference between the infinite sum of discounted rewards for the sequence of nodes chosen by the optimal policy, and for the sequence of nodes given by following our policy for $D$ actions and switching to the optimal policy from then on. Consider the $t^{th}$ path exploration. Let us write $n_{D,t}$ the node that we have after following our policy for $D$ actions. It may be different from the node $n^*_{D}$ that we would have had with the optimal policy. For this reason, $n^*_{D+1}$ may not be available after $n_{D,t}$, which implies that the sequences of nodes that follow can be different, even though we are using the same, optimal policy. Consequently, the immediate regret considered by \citeauthor{Bubeck2010b} is equal to our regret $r_t$, measured up to depth $D$, plus $\gamma^D \sum_{i=1}^{+\infty} 2 \gamma^{i-1}$, where the intermediate reward differences after $D$ actions are all bounded by $2$ (since rewards lie in $[-1,1]$). Thus, stopping the exploration at depth $D$ implies a cost in the order of $\gamma^D$ on the simple regret. Clearly, fixing $D$ implies a simple regret in $O(1)$ and is a poor choice. It is necessary to go deeper down the tree as the number of iterations $T$ -- fixed in advance -- increases. This is why OLOP builds a tree which depth depends on $T$.
		
		GPTS can also be used in a similar fashion, by fixing $T$ and choosing $D$ as a function of $T$. This makes $N$ depend on $T$, and $I^*(T)$ will not be bounded by a constant anymore. We adapt our previous results by determining $\bl_i$ for the current kernel. If two paths differ on $j$ nodes, they have $D(T)-j$ action indices in common.
		\begin{eqnarray*}
			\chi_j - \chi_{j+1} & = & \frac{(\gamma^2)^{D(T)-(j+1)} - (\gamma^2)^{D(T)-j}}{1-\gamma^2}\\
			& = & (\gamma^2)^{D(T)-j-1}\\
			\bl_i & = & \sum_{j=0}^{i-1} B^j (\chi_j-\chi_{j+1})\\
			& = & \frac{\gamma^{2 D(T)}}{\gamma} \sum_{j=0}^{i-1} (\frac{B}{\gamma^2})^j
		\end{eqnarray*}
		This expression is very similar to the one obtained for the linear kernel, but with $\frac{B}{\gamma^2}$ in place of $B$ and $\frac{\gamma^{2 D(T)}}{\gamma}$ in place of $\frac{1}{D+1}$. As a result, $\hat{L}(t)$ can be taken as $\frac{\gamma^{2 D(T)} N(T) B}{(B-\gamma^2) \gamma t}$ which implies $I^*(T) = \tO(\gamma^{2 D(T)}\ N(T))$. Taking $D(T) = \log_B(T)$ -- as OLOP does -- implies $N(T) = T$ and $I^*(T) = \tO(T^{1-2\log_B(\frac{1}{\gamma})})$. The regret becomes:
		\begin{eqnarray*}
			R_T & = & O(\sqrt{T I^*(T) \log(N(T))}) + O(T \gamma^{D(T)})\\
			%& = & \tO(\sqrt{T^{2-2\log_B(\frac{1}{\gamma})} \log(T)}) + O(T^{1-\log_B(\frac{1}{\gamma})})\\
			& = & \tO(T^{1-\log_B(\frac{1}{\gamma})})
		\end{eqnarray*}
		with high probability. This is similar to the OLOP bound for the case where $\gamma^2 B > 1$, but with $T$ instead of $n$. We write $a = \log_B(\frac{1}{\gamma}) > 0$. We have that $1 - a > 0$ since $\gamma B > 1$. Using $n = D\ T = T \log_B(T)$ we can show that the OLOP bound in $n$ implies the same bound as ours in $T$.
		\begin{eqnarray*}
			\frac{1}{n} R_n^{(\mathrm{olop})} & = & \tO(n^{-a})\\
			\exists \alpha, \beta > 0, R_n^{(\mathrm{olop})} & \leq & \alpha \log(n^{-a})^\beta n^{1-a} \mbox{ where } \beta \mbox{ is even}\\
			& \leq & \frac{(-a)^\beta}{(1-a)^\beta} \alpha \log(n^{1-a})^\beta n^{1-a}\\
			R_n^{(\mathrm{olop})} & = & \tO(n^{1-a})\\
			& \leq & \alpha \log(n^{1-a})^{\beta} n^{1-a}\\
			& \leq & \alpha (1-a) \log(n)^{\beta} T^{1-a} \log_B(T)^{1-a}\\
			& \leq & \alpha' \log(T)^{\beta+1-a} T^{1-a}\\
			& \leq & \frac{\alpha'}{(1-a)^{\beta+1-a}} \log(T^{1-a})^{\beta+1-a} T^{1-a}\\
			& = & \tO(T^{1-a})
		\end{eqnarray*}
		
		Note that the simple regret bound of GPTS in $\tO(T^{-\log_B(\frac{1}{\gamma})})$ is better than that of OLOP in $\tO(T^{-\frac{1}{2}})$ when $\frac{1}{B} < \gamma \leq \frac{1}{\sqrt{B}}$, which is understandable since our assumptions are stronger.

\subsection{Possible extensions: a few ideas}

The ideas introduced in this work can be further developed and extended to some particular tree search problems, as mentioned in this section.

	\subsubsection{Variants with different outputs and stopping criteria}
		We mentioned in the introduction that we could use the confidence intervals built by GPB to change the output of our algorithm: instead of outputting the best observed path, we could output the path with highest lower confidence bound for instance. When outputting an optimal action to take from the root, in the MDP planning case, we could output the action that has highest estimated reward in the long term. These variants might be more robust to the variability of the rewards (we could be misled by an unlikely high reward value for a mediocre path), but we do not have regret bounds for them. Confidence intervals can also be used to determine a stopping criterion, e.g. stop when the width of the confidence interval (at a given confidence threshold) for the best observed action is smaller than a certain threshold. We then automatically have a performance guarantee for our algorithm, but no guarantee on the runtime.

	\subsubsection{Hierarchical optimisation}
		We can use GPTS in a manner similar to HOO, with $B=2$ and $D = O(\log(T))$, to find the maximum of a function in a space for which we are given a tree of coverings. Each leaf node of the tree corresponds to a region of the search space (these get smaller as $D$ increases), and we aim at learning the average $f$ values in these regions. The search space can be the Cartesian product of an arbitrary (and possibly infinite) family of discrete and continuous sets. Hierarchical Optimistic Optimisation has the advantage that the choice of a point to sample the function at is very straightforward, it doesn't require any heuristics as in GP Optimisation, and it offers a unified framework for many optimisation problems (not just in $\Re^d$).

			% If we use a kernel between siblings as described in the previous paragraph, we could even share information between siblings, i.e. between disjoint parts of the input space, which HOO doesn't do but would be useful if the function to optimise is globally smooth.

	\subsubsection{Modelling dependencies between nodes}
		One of the most interesting lines of research for Upper Confidence-type Tree Search algorithms is to generalise between nodes of the tree, according to \citet{Gelly2007}. Domain knowledge can be incorporated in GPB by encoding heuristics in the prior mean, but also by labelling nodes and incorporating a kernel between nodes in our kernel between tree paths.
		
		\paragraph{Go tree search}
			When searching Go game trees, we could simply label nodes by the corresponding Go boards. We would then use a kernel between Go boards, applied to the leaves of two paths. Nodes would be selected in a sequence, starting from the root and computing the upper confidence values of each possible next Go board in order to select the next node. The same instance of the GPB algorithm can be used at each time we need to search for an optimal move, and the knowledge gained by the algorithm can be transferred from one game to the other.
		
		\paragraph{Trees with labelled nodes}
			Nodes can be labelled by feature vectors, and a natural kernel between sequences of nodes would be a product of Gaussian kernels between the feature vectors at same depth: this is the same as creating a feature representation of the path by concatenating the feature vectors of its nodes and then taking a Gaussian kernel in the paths feature space. Let us write $\tilde{K}$ the kernel matrix between the children of a node (assuming this matrix is the same for all non-leaf node of the tree). The method of our eigenanalysis of $K_{B,D}$ could be adapted by first writing $K_{B,D} = \tilde{\BK}(K_{B,D-1})$ where $\tilde{\BK}$ is the $B \times B$ block matrix with coefficients taken from $\tilde{K}$ (itself a $B \times B$ matrix). Here, $\tilde{K}$ takes the role of $J_B$ and $\tilde{\BK}$ takes the role of $\BJ^{(B)}$. However, with regard to the implementation of the algorithm, it is not clear how the search for the maximum of the upper confidence function would be performed.
		
		\paragraph{Planning in MDPs}
			When generative models are available, we could aim to learn immediate reward functions at each node, as functions of the children's labels. There would be one GPB instance per node, and after the exploration of a path we would train each instance for each node along that path with the corresponding immediate reward that was observed (this assumes that we can observe immediate rewards and not only the discounted sum of these rewards). The selection of nodes would be performed in a way similar to UCT, by using a sequence of UCB-type bandit algorithms. We can get simple regret bounds at each node $n$, for $f_n$ being the immediate mean-reward function at this node, that take advantage of the spectral properties of $\tilde{K}$. We can then combine these to get simple regret bounds for whole paths and $f$ being the discounted sum of immediate mean-rewards.
				% 
				% Otherwise, we could adopt the exact same strategy as UCT by using a sequence of GPB algorithms (one per node) in order to select nodes to visit. Each GPB algorithm would be equipped with a $B \times B$ kernel matrix. However, although conceptually simple, this could prove difficult to analyse. UCT aims to learn, for each node, the average reward value for all paths going through this node. Because the following nodes are chosen by bandit algorithms, the rewards that are back-propagated are not iid. For the same reason, the sequence of noise variables for each GPB algorithm will not be a martingale sequence (a bias is introduced by the other algorithms further down the tree).

	\subsubsection{Closed-loop planning in MDPs}
		Finally, this work might be extended to closed-loop planning in communicating MDPs with deterministic transitions, by considering cycles through the graph of states instead of paths through a tree of given depth: all cycles have a length smaller than the diameter $D$ of the MDP, which is finite in communicating MDPs.
		% regret analysis for GPB with fixed depth equal to MDP diameter
		% based on a problem-specific measure such as a proportion of short cycles (we prefer cycles to be all of same length, so that our model fits better) -> future work
		Closed-loop planning differs from open-loop in the fact that the chosen actions depend on the current states and not only on time. If no generative model of the MDP is available, we would directly interact with the environment and we would therefore use the cumulative regret as a measure of performance, since every interaction with the environment would have a cost. If there are dependencies between actions, GP inference could be used to derive tighter upper confidence bounds to be used in an algorithm such as UCycle \citep{Ortner2010}.
		% Another idea would be to learn the immediate reward function with a GPB algorithm by considering state-action couples as arms of a bandit problem ?

\subsection{Conclusion and future work}
	
	To sum up, in this paper we have presented a bandit-based Tree Search algorithm which makes use of the Gaussian Processes framework to model the reward function on leaves. The resulting assumption on the smoothness of the function is easily configurable through the use of covariance functions -- which parameters can be learnt by maximum likelihood if not known in advance. We have analysed the regret of the algorithm and provided problem-independent bounds with tight constants, expressed in terms of the $\chi_{d, 0 \leq d \leq D}$ parameters of the covariance function between paths. When comparing to other algorithms, we have shown in particular that GPTS applied to planning in MDPs achieves same regret growth rate as the recent OLOP algorithm.
	
	We believe that the analysis presented in this paper will provide groundwork for studying the theoretical properties of some of the extensions mentioned previously. It should also be possible to extend our results to the more agnostic setting where $f$ is a function with finite norm in a given RKHS, by using another bound derived by \citet{Srinivas2010} -- although that bound involves a $I^*(T)$ term instead of $\sqrt{I^*(T)}$, and it is not known yet whether it is optimal. It may also be of interest to derive bounds for non noisy observations of $f$ (for planning in deterministic environments, for instance), and to analyse the number of times we play sub-optimal arms, in order to get problem-specific bounds \citep[as done by][]{Audibert2007}. Finally, to complement the theoretical analysis of the algorithm, we should investigate the performance of GPB on practical Tree Search problems. In particular, it would most certainly be interesting to see if the use of a kernel between Go boards could be beneficial compared to other techniques used in Go AI such as UCT-RAVE.

		% compare with LinRel for Tree Search (OLOP compared to several many-armed bandit algorithms but not this one)

% \subsection{Comparison with iterative deepening}
% unlike iterative deepening, we force the alg to go to the bottom of the tree (depth D) at each iteration
% but it is creating dummies along the way
% 
% -> we can `jump' immediately to a depth d if algorithm chooses dummy at depth d
% we're then doing a random walk and adding (D-d) nodes and dummies to the memory

%\section{Conclusion}
%\label{sec:conclusion}
%\input{conclusion}

\bibliography{/Users/luigi/Work/Dropbox/PhD/library}

\end{document}